\documentclass[10pt,twocolumn,letterpaper]{article}

\usepackage[pagenumbers]{wacv}     

\usepackage{graphicx}
\usepackage{amsmath}
\usepackage{amssymb}
\usepackage{booktabs}

\usepackage{url}            
\usepackage{amsfonts}       
\usepackage{nicefrac}       
\usepackage{microtype}      
\usepackage{xcolor}         

\usepackage{times}
\usepackage{epsfig}
\usepackage{multirow}
\usepackage{pifont}
\usepackage{diagbox}
\usepackage{bm}

\usepackage{float}

\usepackage[pagebackref,breaklinks,colorlinks]{hyperref}

\usepackage[capitalize]{cleveref}
\crefname{section}{Sec.}{Secs.}
\Crefname{section}{Section}{Sections}
\Crefname{table}{Table}{Tables}
\crefname{table}{Tab.}{Tabs.}

\begin{document}

\title{Universal Test-time Adaptation through Weight Ensembling, Diversity Weighting, and Prior Correction}

\author{%
Robert A. Marsden \thanks{Equal contribution.} \qquad  Mario D\"obler \footnotemark[1] \qquad Bin Yang \\
University of Stuttgart\\
{\tt\small  \{robert.marsden, mario.doebler, bin.yang\}@iss.uni-stuttgart.de}
}

\maketitle

\begin{abstract}
Since distribution shifts are likely to occur during test-time and can drastically decrease the model's performance, online test-time adaptation (TTA) continues to update the model after deployment, leveraging the current test data. Clearly, a method proposed for online TTA has to perform well for all kinds of environmental conditions. By introducing the variable factors domain non-stationarity and temporal correlation, we first unfold all practically relevant settings and define the entity as universal TTA. We want to highlight that this is the first work that covers such a broad spectrum, which is indispensable for the use in practice. To tackle the problem of universal TTA, we identify and highlight several challenges a self-training based method has to deal with: 1) model bias and the occurrence of trivial solutions when performing entropy minimization on varying sequence lengths with and without multiple domain shifts, 2)~loss of generalization which exacerbates the adaptation to multiple domain shifts and the occurrence of catastrophic forgetting, and 3) performance degradation due to shifts in class prior. To prevent the model from becoming biased, we leverage a dataset and model-agnostic certainty and diversity weighting. In order to maintain generalization and prevent catastrophic forgetting, we propose to continually weight-average the source and adapted model. To compensate for disparities in the class prior during test-time, we propose an adaptive prior correction scheme that reweights the model's predictions. We evaluate our approach, named ROID, on a wide range of settings, datasets, and models, setting new standards in the field of universal TTA. Code is available at: \url{https://github.com/mariodoebler/test-time-adaptation}.
\end{abstract}


\section{Introduction}
Deep neural networks achieve remarkable performance, as long as training and test data originate from the same distribution. However, in the real world, environmental changes can occur during test-time and will likely degrade the performance of the deployed model. Domain generalization aims to address potential domain shifts by improving the robustness and generalization of the model directly during training \cite{hendrycks2019augmix, hendrycks2021many, muandet2013domain, tobin2017domain, tremblay2018training}. Due to the wide range of data shifts \cite{quinonero2008dataset} which are typically unknown during training \cite{mintun2021interaction}, the effectiveness of these approaches remains limited. Since the test data provide insights into the current distribution shift, online test-time adaptation (TTA) emerged. In TTA, the model is adapted directly during test-time using an unsupervised loss function like the entropy and the available test sample(s) at time step $t$. 

\begin{figure*}[t]
\vspace{-0.15in}
\centering
\scalebox{0.72}{
\def\svgwidth{580pt}
\graphicspath{{figures/}}
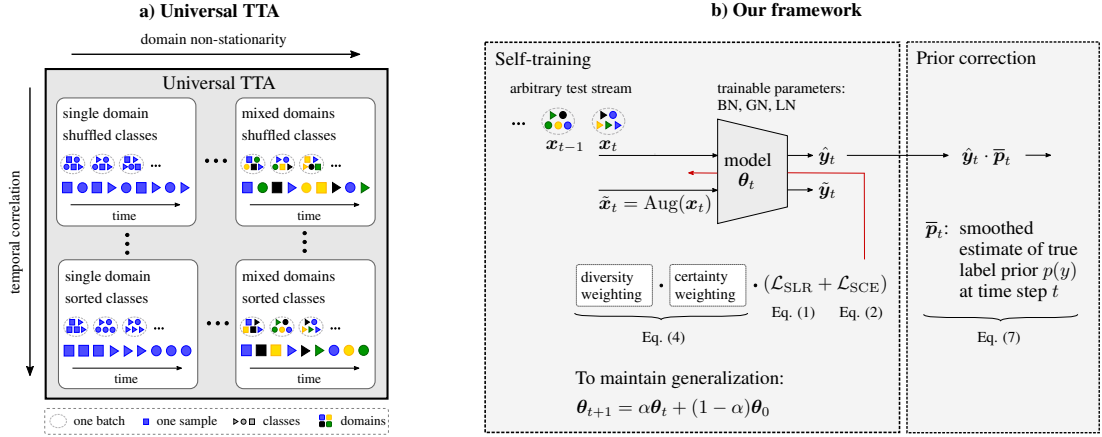
}
\vspace{-0.05in}
\caption{Illustration of universal TTA for a single or a batch of test samples and our framework ROID.}
\label{fig:settings}
\vspace{-0.15in}
\end{figure*}

Although TENT \cite{wang2021tent} has demonstrated success in adapting to single domain shifts, recent research on TTA has identified more challenging scenarios where methods solely based on self-training, such as TENT, often fail \cite{wang2022continual, niu2023towards, gong2022note, yuan2023robust, boudiaf2022parameter}. However, these studies again have predominantly focused on specific settings, overlooking the broad spectrum of possible scenarios. Therefore, we initiate our approach by identifying two key factors that encompass all practically relevant scenarios: \textit{domain non-stationarity} and \textit{temporal correlation}. We denote the complete set of scenarios, including the capability to adapt to arbitrary domains, as \textit{universal TTA}, illustrated in \Cref{fig:settings} a).

In the following, we highlight the challenges imposed by these environmental factors and derive design choices for our framework ROID. Starting with the simplest scenario of adapting to a single domain with i.i.d. data, we empirically show that even when encountering a uniform class distribution a self-training based approach is likely to develop a bias towards certain classes. This poses the risk that when adapting to long sequences, a model collapse is likely, where finally only a small subset or a single class is predicted. Therefore, maintaining diverse predictions is essential. To address this, we introduce a dataset and model-agnostic certainty and diversity loss weighting.

Considering the degree of \textit{domain non-stationarity}, common scenarios range from gradual or continual domain shifts \cite{wang2022continual, GTTA} to consecutive test samples originating from different domains. To deal with non-stationarity, maintaining diversity is even more crucial. We empirically show that the presence of multiple domain shifts can explicitly trigger a collapse to a trivial solution. In contrast to the single domain scenario, continual TTA \cite{wang2022continual} considers the adaptation to a sequence of multiple domains. In this context, in order to ensure effective adaptation to future shifts, a model must uphold its generalization. We hypothesize that adapting a model through self-training on a narrow distribution deteriorates generalization. This is validated by our empirical observations, indicating that a stronger adaptation results in a higher generalization error and promotes catastrophic forgetting. In response, we propose to continually weight-average the current model with the initial source model and denote this as weight ensembling. Dealing with mixed domains presents additional difficulties, such as adapting to multiple target domains simultaneously and the ineffectiveness of covariate shift mitigation through recalculating the batch normalization (BN) statistics during test-time \cite{schneider2020improving}.

In case of \textit{temporally correlated} data or single sample TTA, the estimation of reliable BN statistics is not possible. While introducing a buffer can mitigate this problem \cite{yuan2023robust}, it can raise privacy and memory issues. Alternatively, one can leverage normalization layers like group normalization (GN) or layer normalization (LN), which do not require a batch of data to estimate the statistic and are thus better suited \cite{TTT, niu2023towards}. Since applying diversity weighting promotes the model output to be unbiased, i.e., approximately uniform, even a model that is well adapted to the current domain shift will underperform in a temporally correlated setting. This is due to the existing shift in the class prior. Therefore, instead of allowing the model to become biased, we propose prior correction which introduces an adaptive additive smoothing scheme to reweight the model's predictions.

We summarize our contributions as follows: 1) Our proposed method significantly outperforms existing approaches in the challenging setting of universal TTA. This indicates the potential of our method to be used in practical scenarios. 2) Through our analysis, we provide valuable insights into the challenges that arise when models are subjected to self-training during test-time. 3) Depending on the application, single-sample TTA might be of interest. We highlight that architectures that do not rely on batch normalization layers allow to recover the batch TTA setting from a single sample scenario by doing gradient accumulation. This also dramatically reduces memory consumption. 4) We show that current methods, even if proposed for challenging settings, often fail to fully address the whole picture of universal TTA---a result of our extensive and broad experiments in terms of settings, domain shifts, and models.


\section{Related Work}
\textbf{Unsupervised domain adaptation}
Since domain generalization has its limitations due to the high amount of possible domain shifts that are unknown during training, in the field of unsupervised domain adaptation (UDA) \cite{wilson2020survey}, labeled source and unlabeled target data are used to adapt to the target domain. One line of work minimizes the discrepancy between the source and target feature distribution by either using adversarial learning \cite{ganin2016domain, tzeng2017adversarial}, discrepancy based loss functions \cite{chen2020homm, sun2016deep, yan2017mind}, or contrastive learning \cite{kang2019contrastive, marsden2022contrastive}. Instead of aligning the feature space, it is also possible to align the input space \cite{sankaranarayanan2018generate, CYCADA, ACE, marsden2022continual}, e.g., via style-transfer. Recently, self-training based approaches have shown to be powerful. Self-training uses the networks' predictions on the target domain as pseudo-labels to minimize, e.g., a (cross-)entropy loss \cite{IAST, ADVENT, liu2021cycle, CRST}. Often filtering pseudo-labels is applied to remove unreliable samples. Mean teachers \cite{tarvainen2017mean} can be further leveraged to increase the reliability of the network's predictions \cite{tranheden2021dacs, french2018selfensembling}. 

\textbf{Test-time adaptation}
While UDA typically performs offline model adaptation, online test-time adaptation adapts the model to an unknown domain shift directly during inference using the currently available test samples. \cite{schneider2020improving} showed that estimating new batch normalization (BN) statistics during test-time can significantly improve the performance on shifts caused by corruptions. While only updating the BN statistics is computationally efficient, it has its limitations, especially when it comes to natural domain shifts. Therefore, recent TTA methods further update the model weights by relying on self-training. TENT \cite{wang2021tent} demonstrated that minimizing the entropy with respect to the batch normalization parameters can be successful for single-target adaptation. EATA \cite{niu2022efficient} extends this idea by weighting the samples according to their reliability and diversity. Further, they use elastic weight consolidation \cite{kirkpatrick2017overcoming} to prevent catastrophic forgetting \cite{mccloskey1989catastrophic} on the initial training domain. However, this requires access to data from the initial training domain, which is not always available in practice. To circumvent a model collapse to trivial solutions caused by confidence maximization, \cite{liang2020we, mummadi2021test} make use of diversity regularizers. Contrastive learning has also found its application in TTA \cite{chen2022contrastive, dobler2023robust}.

While some TTA methods only consider the adaptation to a single domain, in the real world, it is common to encounter multiple domain shifts. Therefore, \cite{wang2022continual} introduced the setting of continual test-time adaptation, where a model has to adapt to a sequence of different domains. While self-training based methods such as \cite{wang2021tent} can be applied to the continual setting, they can be prone to error accumulation \cite{wang2022continual}. To prevent error accumulation, \cite{wang2022continual} proposes to use weight and augmentation-averaged predictions in combination with a stochastic restore to mitigate catastrophic forgetting. RMT \cite{dobler2023robust} proposes a robust mean teacher to deal with multiple domain shifts and GTTA \cite{GTTA} uses mixup and style-transfer to artificially create intermediate domains. LAME \cite{boudiaf2022parameter}, NOTE \cite{gong2022note}, SAR \cite{niu2023towards}, and RoTTA \cite{yuan2023robust} propose methods that focus on dealing with temporally correlated data. While LAME only adapts the model's output with Laplacian adjusted maximum-likelihood estimation, NOTE and RoTTA introduce a buffer to simulate an i.i.d. stream. SAR proposes a sharpness-aware and reliable entropy minimization method to be robust to large and noisy gradients.

Further areas of test-time adaptation focus on settings where the collection of a batch of data may not be feasible due to timeliness. Methods for single-sample TTA \cite{mirza2022norm, gao2022back, zhang2021memo, MT3} often rely on artificially creating a batch of data through test-time augmentation \cite{krizhevsky2009learning}, which drastically increases the computational overhead. Due to only using a single sample for adapting the model, updates can be noisy and therefore the adaptation capability may be limited. Further, the area of test-time training modifies the initial pre-training phase by introducing an additional self-supervision loss that is also exploited to adapt the model during test-time \cite{TTT, liu2021ttt++, MT3}. Thus, test-time training is unable to use any off-the-shelf pre-trained model.


\section{Self-training for Test-time Adaptation}
\label{sec:self_training_analysis}
Let $\bm{\theta}_0$ denote the weights of a deep neural network pre-trained on labeled source data $(\mathcal{X}, \mathcal{Y})$. While the network will typically perform well on data originating from the same domain, this is usually not the case when the model encounters data from different domains. This lack of generalization to out of distribution data is a problem in practice since the environmental conditions are likely to change from time to time. To keep the networks' performance high during inference, online test-time adaptation continues to update the model after deployment using an unsupervised loss function like the entropy and the currently available test data $\bm{x}_t$ at time step $t$. 

\begin{figure*}[t]
    \centering
    \vskip -0.1in
    \includegraphics[width=0.85\textwidth]{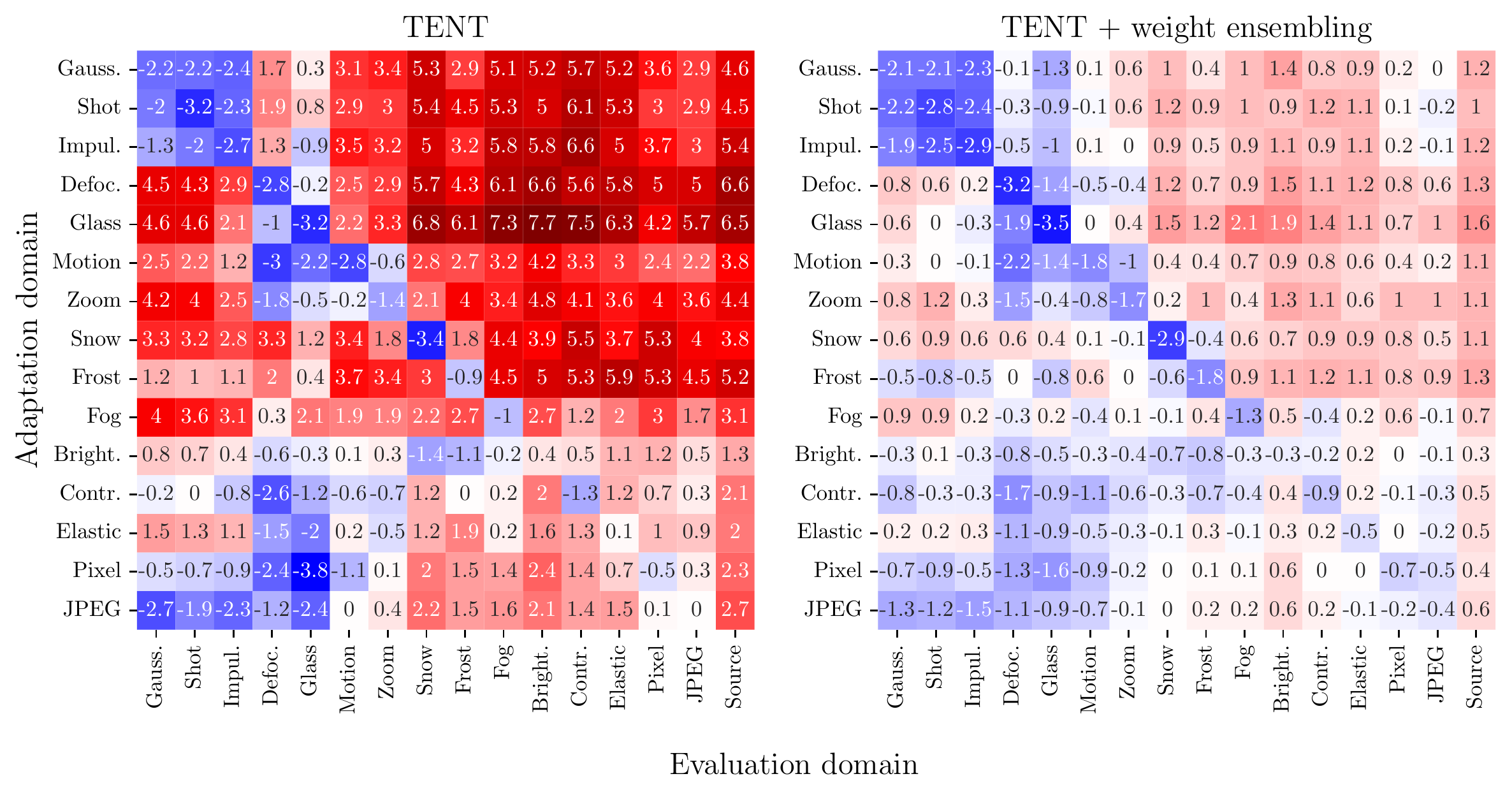}
    \vskip -0.1in
    \caption{Difference of error for a moderate and a stronger adaptation, corresponding to a learning rate of $10^{-4}$ and $10^{-3}$, respectively. An ImageNet pre-trained ResNet-50 is adapted on one of the corruptions from ImageNet-C at severity 3 and evaluated on all corruptions and the source domain.}
    \label{fig:generalization}
    \vskip -0.1in
\end{figure*}
\textbf{Test-time adaption through self-training carries the risk of generalization loss} Adapting a model to a target domain effectively means moving the model from its initial source parameterization to a parameterization that better models the current target distribution. This carries the risk that predictions on the source distribution become inaccurate, but also carries the risk of losing generalization when the target distribution is narrow. The former is known as catastrophic forgetting. We now want to highlight the latter, since generalization is a so far underestimated topic in TTA and is important for coping with non-stationary domains.

To study the impact of performing entropy minimization on generalization, we consider a typical TTA framework (TENT) where only parameters of the BN layers are trained while the rest remains frozen. We utilize an ImageNet pre-trained ResNet-50 and adapt the model using 40,000 samples of one of the corruptions from ImageNet-C \cite{hendrycks2019benchmarking}. To investigate the adaptation and generalization, we then evaluate the adapted model for each corruption on the remaining 10,000 samples. In \Cref{fig:generalization}, we illustrate the difference of error for a moderate and a stronger adaptation, corresponding to a learning rate of $10^{-4}$ and $10^{-3}$, respectively. As one would expect, a stronger adaptation leads to an improvement for samples originating from the same or a similar domain. However, this comes with the drawback that the performance on other domains deteriorates, indicating a loss of generalization. As a result, adapting to future domains is hindered. The same effect can be observed for the source domain, depicted in the last column, showing signs of catastrophic forgetting. As illustrated in \Cref{fig:generalization_appendix} located in \Cref{sec:appendix_generalization}, the effect also occurs for supervised fine-tuning. Using weight ensembling, as described in \Cref{sec:weight_ensembling} and depicted in \Cref{fig:generalization}, retains generalization, while still enabling a good adaptation. \begin{figure}[t]
    \centering
    \includegraphics[scale=0.55]{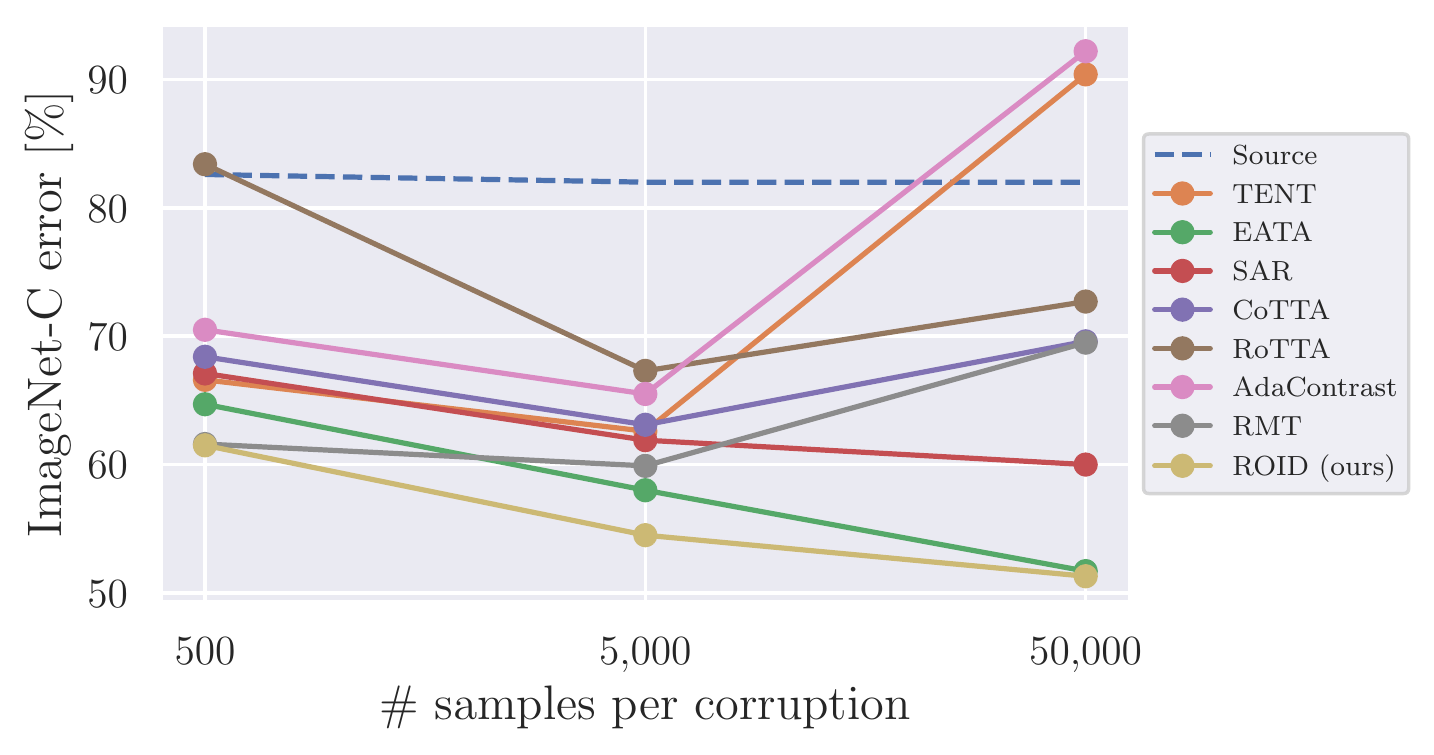}
    \vskip -0.1in
    \caption{Average online classification error rate~(\%) over 5 runs for the task of \textit{continual} TTA on the highest corruption severity level 5 of ImageNet-C. Varying sequence lengths are considered for a ResNet-50.}
    \label{fig:num_ex}
    \vskip -0.2in
\end{figure}

A similar effect was found by \cite{radford2021learning}, who reported that when fine-tuning their zero-shot model CLIP on ImageNet, the model generalization decreases while the performance on the adaptation domain drastically increases. We argue that such a phenomenon is likely to occur to any model that is fine-tuned on a less diverse dataset compared to the initial training dataset. (In case of CLIP, the initial training dataset consists of 400 million images which is approximately 312 times bigger than ImageNet). 

\textbf{Stability} Undoubtedly, the most critical aspect for a successful universal TTA is stability. Although TENT \cite{wang2021tent} has demonstrated a successful adaptation to a single domain shift at a time, we empirically show in \Cref{sec:appendix_error_accumulation} that its performance on ImageNet-C degrades to a trivial solution as the length of the test sequence increases. In addition, by considering CIFAR100-C, we also demonstrate that the occurrence of trivial solutions can be triggered when the domain shifts from time to time---a setting which is likely to be encountered in real world applications and denoted as continual TTA by \cite{wang2022continual}. We further find that an increased domain non-stationarity has an even more severe effect, as the model develops a bias much faster. In \Cref{fig:num_ex}, we analyze the performance of current state-of-the-art methods in the online continual TTA setting for ImageNet-C, using different numbers of samples per corruption. While all methods successfully reduce the error rate for 5,000 samples per corruption, only very few methods do not collapse to trivial solutions or again degrade in performance due to the development of a bias when 50,000 samples are considered. We visualize and discuss the latter two aspects in \Cref{sec:appendix_error_accumulation}. These examples clearly demonstrate the necessity of remaining diverse predictions throughout the adaptation.


\section{Methodology}
In this work, we seek to create a method that performs a good, stable, and efficient adaptation across a wide range of different settings and domain shifts while being mostly model agnostic. Before we address the previous findings in more detail, we first establish the basic framework.

To ensure efficiency during test-time, we only update the network's normalization parameters (BN, GN, and LN) and freeze all others. To improve the stability and adaptation, we exchange the commonly used entropy loss by a certainty and diversity weighted version of the soft likelihood ratio (SLR) loss. The SLR loss \cite{mummadi2021test} has the advantage that its gradients are less dominated by low confidence predictions, which are typically more likely to be incorrect \cite{mummadi2021test}. The weighted soft likelihood ratio loss is then given by
\begin{equation}
    \mathcal{L}_\mathrm{SLR}(\bm{\hat{y}}_{ti})=-\sum_{c} w_{ti}\, \hat{y}_{tic}\log(\frac{\hat{y}_{tic}}{\sum_{j\neq c}\hat{y}_{tij}}),
\end{equation}
where $\bm{\hat{y}}_{ti}$ are the softmax probabilities of the network for the $i$-th test sample at time step $t$ and $w_{ti}$ is its corresponding weight. Since the SLR loss encourages to scale the networks' logits larger and larger \cite{mummadi2021test}, we propose to clip the softmax probabilities for very high confidence values, i.e., $\bm{\hat{y}}_t\in [0, 0.99]^C$, where $C$ is the number of classes. This results in a zero-gradient for probabilities above the clipping value, preventing logit explosion.

To further strengthen the adaptation, we encourage consistency against smaller perturbations. This is achieved by promoting similar outputs between test images which have been identified as certain and diverse ($\bm{x}'_{t}$) and an augmented view of them. We use color jitter, affine transformations, and horizontal flipping to generate the augmented view $\bm{\tilde{x}}'_{t} = \mathrm{Aug}(\bm{x}'_{t})$ with predictions $\bm{\tilde{y}}'_{t}$. Subsequently, a weighted consistency loss based on the symmetric cross-entropy (SCE) is calculated
\begin{equation}
    \mathcal{L}_{\mathrm{SCE}}(\bm{\hat{y}}'_{ti}, \bm{\tilde{y}}'_{ti})=-\frac{w'_{ti}}{2} \bigg( \sum_{c=1}^C \hat{y}'_{tic}\, \mathrm{log}\, \tilde{y}'_{tic} + \sum_{c=1}^C \tilde{y}'_{tic}\, \mathrm{log}\, \hat{y}'_{tic}\bigg).
\end{equation}
We leverage the SCE loss due to its tolerance towards label noise \cite{wang2019symmetric}, which is especially important in the setting of self-training where pseudo-labels can be inaccurate.

\subsection{Certainty and diversity weighting}
\label{sec:weighting}
Our analysis in \Cref{sec:self_training_analysis} and \Cref{sec:appendix_error_accumulation} suggests that it is essential to prevent the model from becoming biased or, worse, collapse to a trivial solution during test-time. Therefore, we introduce a diversity criterion, similar to \cite{niu2022efficient}, which ensures that diverse samples are favored in comparison to samples that are similar to the central tendency of recent model predictions. Unlike \cite{niu2022efficient}, we propose a diversity weighting that does not require dataset-specific hyperparameters. We begin by tracking the recent tendency of a model's prediction with an exponential moving average $\bm{\bar{y}}_{t+1}=\beta\, \bm{\bar{y}}_{t} + \frac{(1-\beta)}{N_b} \sum_i^{N_b}\bm{\hat{y}}_{ti}$, setting $\beta=0.9$. To determine a diversity weight for each test sample $\bm{x}_{ti}$, the cosine similarity between the current model output $\bm{\hat{y}}_{t}$ and the tendency of the recent outputs $\bm{\bar{y}}_{t}$ is computed as follows
\begin{equation}
    w_{\mathrm{div},{ti}}=1-\frac{\bm{\hat{y}}_{ti}^\mathrm{T} \, \bm{\bar{y}}_{t}}{\Vert\bm{\hat{y}}_{ti}\Vert \, \Vert\bm{\bar{y}}_{t}\Vert}.
\end{equation}
This strategy has the advantage that if the model output is uniform, uncertain predictions receive a smaller weight, which prevents the incorporation of errors into the model. However, if the model output is biased towards some classes, uncertain predictions will have a large weight, thus promoting error accumulation. Therefore, we additionally utilize certainty weighting based on the negative entropy $w_{\mathrm{cert},{ti}}=-H(\bm{\hat{y}}_{ti})=\sum_c \hat{y}_{tic}\log \hat{y}_{tic}$. To remove model and data dependencies, such as the model's calibration or the number of classes, we normalize the certainty and diversity weights to be in unit range. To pull apart non-reliable and non-diverse samples from reliable and diverse ones, we take the exponential of the product of diversity and certainty weights, scaled by a temperature $\tau$:
\begin{equation}
    \bm{w}_t=\exp\Big(\frac{\bm{w}_{\mathrm{div},t} \, \bm{w}_{\mathrm{cert},t}}{\tau}\Big).
\end{equation}
To re-emphasize diversity, all weights of samples whose diversity is less than the mean diversity are set to zero, i.e., $w_{ti} = 0$ if $w_{\mathrm{div},{ti}}<\mathrm{mean}(\bm{w}_{\mathrm{div}, t})$.

\subsection{Weight ensembling}
\label{sec:weight_ensembling}
Since our analysis in \Cref{sec:self_training_analysis} revealed that self-training is likely to cause a loss of generalization and catastrophic forgetting, we propose weight ensembling. It averages the weights of the source model which potentially has good generalization capabilities and the adapted model, which typically better models the current distribution. Previous literature supports that weight-averaging two models works, if they remain in the same basin of the loss landscape \cite{frankle2020linear}. This is usually true for models which are fine-tuned from the same pre-trained checkpoint \cite{frankle2020linear, izmailov2018averaging, wortsman2022model}. Specifically, we continually ensemble the weights of the initial source model $\bm{\theta}_0$ and the weights of the current model $\bm{\theta}_t$ at time step $t$ using an exponential moving average of the form
\begin{equation}
    \bm{\theta}_{t+1} = \alpha \, \bm{\theta}_t + (1-\alpha)\bm{\theta}_0,
\end{equation}
where $\alpha$ is a momentum term, balancing adaptation and generalization. Since we only update normalization parameters, the memory overhead for storing source weights is neglectable. The advantages of equipping TENT with our weight ensembling approach, using a momentum term five times larger as the learning rate, are illustrated in \Cref{fig:generalization}. Clearly, the strategy prevents drastic decreases in performance on unseen domains while still allowing good adaptation. By inspecting the last column, it also becomes apparent that catastrophic forgetting is largely mitigated.

\subsection{Prior correction during test-time}
Consider the scenario where no domain shift exists and only the class distributions between the training and test data differ. In this case, a non-adapted model will underperfom because the learned posterior $q(y|x)$ will deviate from the actual posterior $p(y|x)$ due to the shift in priors, i.e., $q(y) \neq p(y)$. However, as shown by \cite{royer2015classifier}, optimal performance can be recovered by correcting the deviation in posterior according to $p(y|x) = q(y|x)\frac{p(y)}{q(y)}$. In the context of online TTA with temporally correlated and thus highly imbalanced data, such performance degradation can easily occur. For example, when the actual class prior is highly dynamic. Since our diversity weighting aims to stabilize model adaptation by preventing the network from learning any biases, there will be a discrepancy between the class priors. Therefore, we propose a prior correction that reweights the final predictions by $\frac{p(y)}{q(y)}$ without influencing the adaptation.

As a result of diversity weighting, we assume a uniform distribution for the learned prior $q(y)$. To determine the actual class prior $p(y)$, we suggest to use the sample mean over the current softmax predictions $\bm{\hat{y}}_{ti}$ as a proxy $\hat{\bm{p}}_t=\frac{1}{N_b} \sum_i^{N_b}\bm{\hat{y}}_{ti}$.
Since only $N_b$ test samples are considered for the estimation of the actual class prior, the resulting estimate will be inaccurate. Therefore, an adaptive additive smoothing scheme is proposed
\begin{equation}
    \bar{\bm{p}}_t=\frac{\hat{\bm{p}}_t+\gamma}{1+\gamma N_c},
\end{equation}
where $N_c$ denotes the number of classes and $\gamma$ is an adaptive smoothing factor that is determined by the ratio $\gamma=\mathrm{max}(1/N_b,1/N_c)/\max_c\hat{\bm{p}}_{tc}$. The idea behind this ratio is that if the class distribution within a batch tends to be uniform, $\gamma\geq1$, a strong smoothing is applied ensuring that no class is favored. If the class distribution is strongly biased towards one class, $\gamma\rightarrow\mathrm{max}(1/N_b,1/N_c)$, minor smoothing is applied. In settings with highly imbalanced data, weighting the network's outputs with a smoothed estimate of the class prior can significantly improve the predictions. Uncertain data points can be corrected by taking class prior information into account, while not degrading performance when a uniform class distribution is present.


\section{Experiments}
\textbf{Datasets}
\label{sec:datasets}
We evaluate our approach for a wide range of different domain shifts, including corruptions and natural shifts. Following \cite{wang2021tent}, we consider the corruption benchmark \cite{hendrycks2019benchmarking} consisting of CIFAR10-C, CIFAR100-C, and ImageNet-C. These datasets include 15 types of corruptions with 5 severity levels applied to the validation and test images of ImageNet (IN) and CIFAR, respectively \cite{krizhevsky2009learning}. For the natural domain shifts, we consider ImageNet-R \cite{hendrycks2021many}, ImageNet-Sketch \cite{wang2019learning}, as well as a variation of ImageNet-D \cite{rusak2022imagenet}, which we denote as ImageNet-D109. While ImageNet-R contains 30,000 examples depicting different renditions of 200 IN classes, ImageNet-Sketch contains 50 sketches for each of the 1,000 IN classes. ImageNet-D is based on DomainNet \cite{peng2019moment}, which contains 6 domain shifts (clipart, infograph, painting, quickdraw, real, sketch), and considers samples that are one of the 164 classes that overlap with ImageNet. For ImageNet-D109, we use all classes that have a one-to-one mapping from DomainNet to ImageNet, resulting in 109 classes. We omit the domain \textit{quickdraw} in our experiments since many examples cannot be attributed to a class \cite{saito2019semi}.

\textbf{Considered settings}
All experiments are performed in the online TTA setting, where the predictions are evaluated immediately. To assess the performance of each method for universal TTA, we consider four different settings. The first is the \textit{continual} benchmark \cite{wang2022continual}, where the model is adapted to a sequence of $K$ different domains $\mathcal{D}$ without knowing when a domain shift occurs, i.e. $[\mathcal{D}_1, \mathcal{D}_2, \dots, \mathcal{D}_K]$. For the corruption datasets, the domain sequence comprises 15~corruptions, each encountered at the highest severity level~5. For ImageNet-R and ImageNet-Sketch there exists only a single domain and for ImageNet-D109 the domains are encountered in alphabetical order. The second setting is denoted as \textit{mixed domains}. Since in this case the test data of all domains are randomly shuffled before the adaptation, consecutive test samples are likely to originate from different domains. Third, we examine a \textit{correlated} setting which is similar to the continual one, since the domains are also encountered sequentially. However, in the correlated setting, the data of each domain is sorted by the class label rather than randomly shuffled, resulting in class imbalanced batches. Finally, we also consider the situation where the domains are mixed and the sequence is temporally correlated. Single domain settings are not explicitly considered since any method that succeeds in the continual setting, will also succeed in the single domain setting.

\textbf{Implementation details}
Following previous work \cite{wang2022continual}, a pre-trained WideResNet-28 (WRN-28) \cite{zagoruyko2016wide} and ResNeXt-29 \cite{xie2017aggregated} is used for CIFAR10-to-CIFAR10-C and CIFAR100-to-CIFAR100-C, respectively. For the ImageNet datasets a source pre-trained ResNet-50, a VisionTransformer \cite{dosovitskiy2020image} in its base version with an input patch size of $16\times16$ (Vit-b-16), and a SwinTransformer \cite{liu2021swin} in its base version (Swin-b) are used. Note that for our method, we additionally ablate 28 pre-trained networks available in PyTorch in \Cref{sec:appendix_architectures}. We follow the implementation of \cite{wang2021tent}, using the same hyperparameters. Further, we fix the momentum term $\alpha$ used for weight ensembling to 0.99 and set the temperature $\tau$ to $\frac{1}{3}$.

\textbf{Baselines}
We compare our approach to other source-free TTA methods that also use an arbitrary off-the-shelf pre-trained model. In particular, we compare to TENT non-episodic \cite{wang2021tent}, EATA \cite{niu2022efficient}, SAR \cite{niu2023towards}, CoTTA \cite{wang2022continual}, RoTTA \cite{yuan2023robust}, AdaContrast \cite{chen2022contrastive}, RMT \cite{dobler2023robust}, and LAME \cite{boudiaf2022parameter}. In addition, we consider the non-adapted model (source) and the normalization-based method BN--1, which recalculates the batch normalization statistics using the current test batch. As metric, we use the error rate.

\subsection{Results}
\label{sec:results_continual}
\begin{table*}[t]
\renewcommand{\arraystretch}{1.2}
\centering
\caption{Average online classification error rate~(\%) over 5 runs in the \textit{continual} TTA setting.} \label{tab:continual}
\vskip -0.1in
\scalebox{0.8}{
\tabcolsep4pt
\begin{tabular}{l|l|cccccccccc|c}
\hline
Dataset & Architecture & Source & BN--1 & TENT & EATA & SAR & CoTTA & RoTTA &  AdaCont. & RMT & LAME & ROID (ours) \\
\hline
CIFAR10-C & WRN-28 & 43.5 & 20.4 & 20.0 & 17.9 & 20.4 & 16.5 & 19.3 & 18.5 & 17.0 & \textcolor{red}{64.3} & \textbf{16.2}$\pm$0.05 \\
CIFAR100-C & ResNext-29 & 46.4 & 35.4 & \textcolor{red}{62.2} & 32.2 & 32.0 & 32.8 & 34.8 & 33.5 & 30.2 & \textcolor{red}{98.5} & \textbf{29.3}$\pm$0.04 \\
\hline
\multirow{3}{*}{IN-C} & ResNet-50 & 82.0 & 68.6 & 62.6 & 58.0 & 61.9 & 63.1 & 67.3 & 65.5 & 59.9 & \textcolor{red}{93.5} & \textbf{54.5}$\pm$0.1 \\
& Swin-b & 64.0 & 64.0 & 64.0 & 52.8 & 63.7 & 59.3 & 62.7 & 58.1 & 52.6 & \textcolor{red}{84.8} & \textbf{47.0}$\pm$0.26 \\
& ViT-b-16 & 60.2 & 60.2 & 54.5 & 49.8 & 51.7 & \textcolor{red}{77.0} & 58.3 & 57.0 & \textcolor{red}{72.9} & \textcolor{red}{79.9} & \textbf{45.0}$\pm$0.09 \\
\hline
\multirow{3}{*}{IN-R} & ResNet-50 & 63.8 & 60.5 & 57.6 & 54.2 & 57.5 & 57.4 & 60.7 & 58.9 & 56.1 & \textcolor{red}{99.3} & \textbf{51.2}$\pm$0.11 \\
& Swin-b & 54.2 & - & 53.8 & 49.9 & 53.0 & 52.9 & 53.0 & 52.3 & 47.4 & \textcolor{red}{92.7} & \textbf{45.8}$\pm$0.12 \\
& ViT-b-16 & 56.0 & - & 53.3 & 49.0 & 48.6 & \textcolor{red}{69.6} & 54.4 & 54.2 & \textcolor{red}{68.8} & \textcolor{red}{95.2} & \textbf{44.2}$\pm$0.13 \\
\hline
\multirow{3}{*}{IN-Sketch} & ResNet-50 & 75.9 & 73.6 & 69.5 & 64.5 & 68.4 & 69.5 & 70.8 & 73.0 & 68.4 & \textcolor{red}{99.8} & \textbf{64.3}$\pm$0.16 \\
& Swin-b & 68.4 & - & \textcolor{red}{68.7} & 60.5 & \textcolor{red}{72.6} & \textcolor{red}{71.0} & 67.1 & 64.4 & \textcolor{red}{69.0} & \textcolor{red}{94.6} & \textbf{58.8}$\pm$0.15 \\
& ViT-b-16 & 70.6 & - & 70.5 & 59.7 & 70.6 & \textcolor{red}{95.5} & 69.0 & 68.3 & \textcolor{red}{86.8} & \textcolor{red}{99.5} & \textbf{58.6}$\pm$0.07 \\
\hline
\multirow{3}{*}{IN-D109} & ResNet-50 & 58.8 & 55.1 & 52.9 & 51.6 & 52.2 & 50.8 & 52.3 & 50.4 & 49.4 & \textcolor{red}{85.0} & \textbf{48.0}$\pm$0.06 \\
& Swin-b & 51.4 & - & \textcolor{red}{66.1} & 47.5 & \textcolor{red}{54.2} & 49.9 & 48.7 & 47.3 & 47.6 & \textcolor{red}{86.3} & \textbf{45.1}$\pm$0.10 \\
& ViT-b-16 & 53.6 & - & \textcolor{red}{84.0} & 47.4 & \textcolor{red}{57.4} & \textcolor{red}{73.4} & 51.2 & 49.7 & \textcolor{red}{74.2} & \textcolor{red}{88.0} & \textbf{45.0}$\pm$0.04 \\
\hline
\end{tabular}
}
\vskip -0.1in
\end{table*}

\begin{table*}
\renewcommand{\arraystretch}{1.2}
\centering
\caption{Average online classification error rate~(\%) over 5 runs in the \textit{mixed domains} TTA setting.} \label{tab:non_stationary}
\vskip -0.1in
\scalebox{0.8}{
\tabcolsep4pt
\begin{tabular}{l|l|cccccccccc|c}\hline
Dataset & Architecture & Source & BN--1 & TENT & EATA & SAR & CoTTA & RoTTA & AdaCont. & RMT & LAME & ROID (ours) \\
\hline
CIFAR10-C & WRN-28 & 43.5 & 33.8 & \textcolor{red}{44.1} & 28.6 & 33.8 & 32.5 & 33.4 & \textbf{26.2} & 31.0 & \textcolor{red}{75.2} & 28.0$\pm$0.12 \\
CIFAR100-C & ResNext-29 & 46.4 & 45.8 & \textcolor{red}{82.5} & 36.9 & 45.5 & 43.1 & 45.4 & 41.8 & 38.6 & \textcolor{red}{98.4} & \textbf{35.0}$\pm$0.04 \\
\hline
\multirow{3}{*}{IN-C} & ResNet-50 & 82.0 & \textcolor{red}{82.5} & \textcolor{red}{86.4} & 72.3 & 79.4 & 76.0 & 78.1 & \textcolor{red}{90.8} & 75.4 & \textcolor{red}{95.1} & \textbf{69.5}$\pm$0.13 \\
& Swin-b & 64.0 & - & 62.6 & 56.3 & 60.6 & 63.3 & 62.6 & \textcolor{red}{66.0} & 55.4 & \textcolor{red}{64.6} & \textbf{55.0}$\pm$0.26 \\
& ViT-b-16 & 60.2 & - & 55.0 & 51.8 & 52.3 & \textcolor{red}{89.3} & 58.2 & \textcolor{red}{65.5} & \textcolor{red}{73.4} & \textcolor{red}{62.6} & \textbf{50.7}$\pm$0.08 \\
\hline
\multirow{3}{*}{IN-D109} & ResNet-50 & 58.8 & 56.2 & 56.1 & 53.3 & 53.7 & \textbf{50.3} & 54.0 & 55.4 & 50.7 & \textcolor{red}{99.1} & 50.9$\pm$0.04 \\
& Swin-b & 51.4 & - & \textcolor{red}{61.5} & 48.9 & \textcolor{red}{54.0} & 49.4 & 48.1 & 49.4 & \textbf{46.5} & \textcolor{red}{97.3} & 47.2$\pm$0.07 \\
& ViT-b-16 & 53.6 & - & \textcolor{red}{76.7} & 48.6 & \textcolor{red}{61.4} & \textcolor{red}{58.0} & 50.5 & 51.4 & \textcolor{red}{70.8} & \textcolor{red}{98.8} & \textbf{46.9}$\pm$0.02 \\
\hline
\end{tabular}
}
\vskip -0.1in
\end{table*}

\textbf{Results for continual TTA} \Cref{tab:continual} shows the results for online continual TTA, with results worse than the source performance highlighted in red. We find that LAME significantly decreases the performance on all continual benchmarks, due to its tendency of predicting only a reduced number of classes in each batch. This can also be seen in \Cref{fig:diversity_pred} in the appendix. While SAR is able to adapt to corrupted data for all architectures, its adaptation capabilities for natural domain shifts are limited when using transformers. Further, although SAR proposed a model restore approach to avoid performance degradation, the approach lacks generalization. The effectiveness of TENT also heavily depends on the domain shift and architecture, as Vit-b-16 provides clear benefits for IN-C and IN-R, but fails for IN-D109, for example. However, by equipping TENT with a diversity criterion, TENT remains stable in all configurations, suggesting that diversity also contributes to become more model and shift agnostic. This might also be the reason, why methods like EATA, AdaContrast and RoTTA remain stable, as each of them either explicitly enforce diversity or leverages a diversity buffer. Our method ROID is not only stable, but yields significant performance improvements compared to the second best approach, EATA, which requires dataset specific hyperparameters and access to data from the initial source domain. Note that we additionally verify the effectiveness of ROID for 28 pre-trained networks in \Cref{sec:appendix_architectures}, demonstrating its wide applicability.

\textbf{Results for mixed domains} \Cref{tab:non_stationary} illustrates the results for the mixed domains setting. By comparing the performance between the settings \textit{continual} and \textit{mixed domains} for methods such as EATA, SAR, AdaContrast, RMT, and ROID for the transformers, it becomes obvious that adapting to multiple target domains at the same time is more challenging. In case of BN-based architectures, like ResNets, the results can also significantly decrease due to missing improvements of covariate shift mitigation through recalculating the BN statistic. Our method ROID is again not only stable, but performs best or comparable on most benchmarks.

\begin{figure*}[t]
    \centering
    \vskip -0.15in
    \begin{tabular}{cc}
        \includegraphics[scale=0.55]{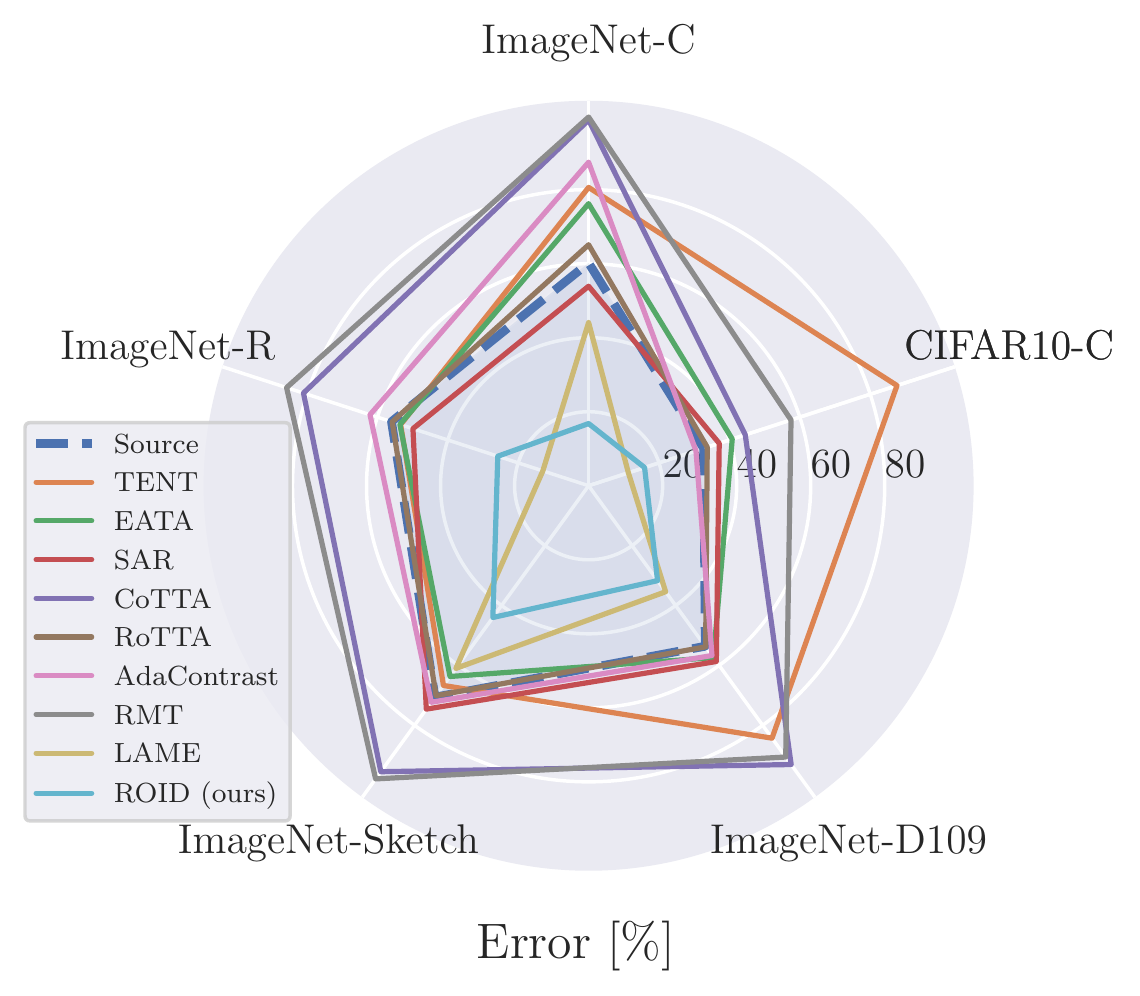} & \includegraphics[scale=0.5]{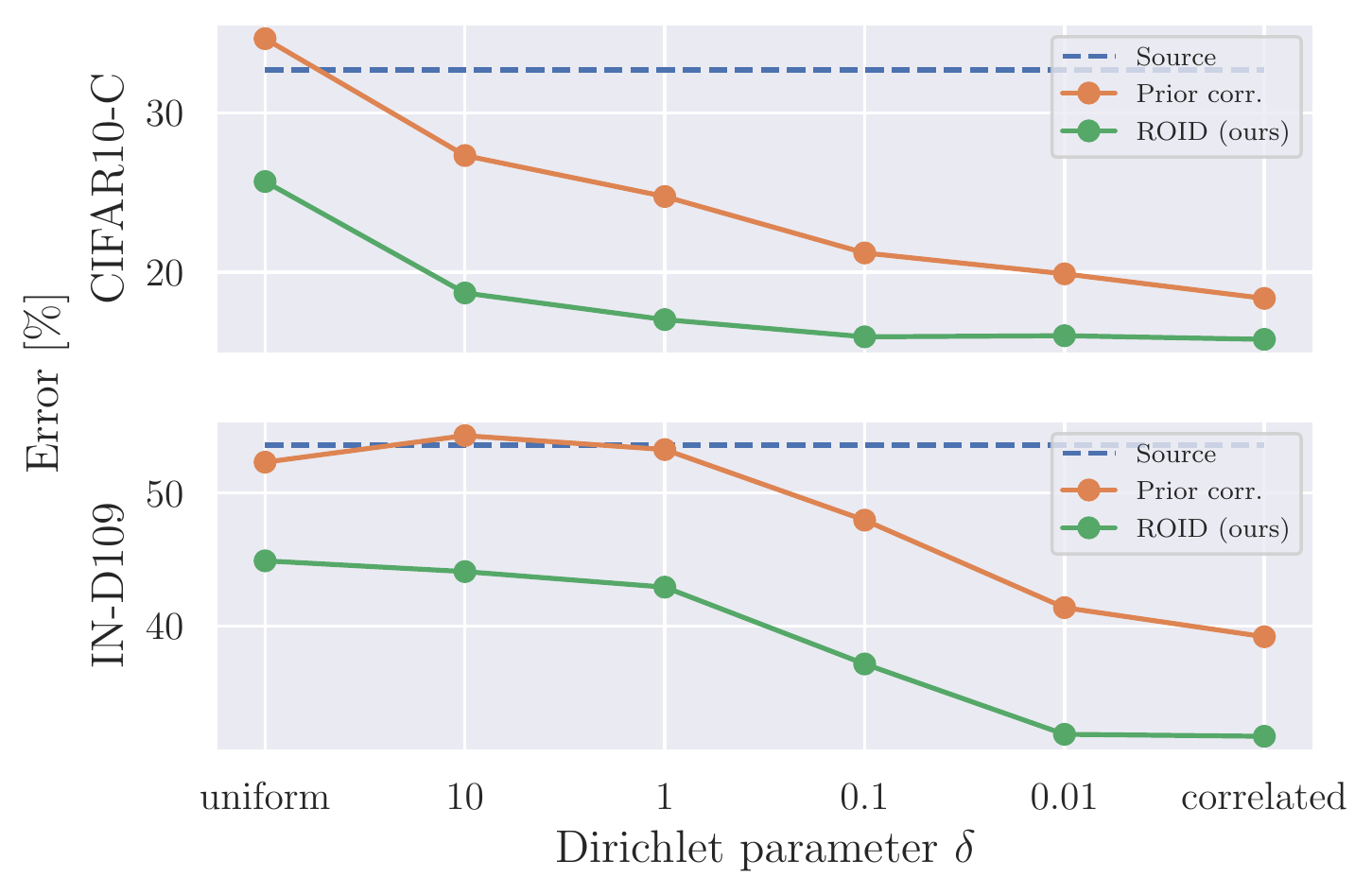} \\
    \end{tabular}
    \vskip -0.15in
    \caption{Online classification error rate~(\%) in the \textit{correlated} TTA setting, where samples are sorted by class on the left and for different levels of correlation on the right.}
    \label{fig:correlated}
    \vskip -0.15in
\end{figure*}
\begin{table}[h]
\renewcommand{\arraystretch}{1.2}
\centering
\caption{Average online classification error rate~(\%) for IN-C (at level 5) and IN-D109 for the \textit{mixed domains correlated} setting,  using $\delta=0.01$ and $\delta=0.1$, respectively.} \label{tab:mixed-correlated}
\vskip -0.1in
\scalebox{0.8}{
\tabcolsep4pt
\begin{tabular}{l|l|c|c}\hline
& Method & IN-C & IN-D109 \\
\hline
\multirow{4}{*}{\rotatebox[origin=c]{90}{Swin-b}} & Source & 64.0 & 51.4 \\
& SAR  & \textcolor{red}{64.9}$\pm$0.81 & \textcolor{red}{53.9}$\pm$0.52 \\
& LAME & 37.4$\pm$0.12 & \textbf{28.0}$\pm$0.39 \\
& ROID & \textbf{28.6}$\pm$0.16 & 28.3$\pm$0.19 \\
\hline
\multirow{4}{*}{\rotatebox[origin=c]{90}{ViT-b-16}} & Source & 60.2 & 53.6 \\
& SAR  & 54.3$\pm$0.59 & \textcolor{red}{60.8}$\pm$0.48 \\
& LAME & 36.1$\pm$0.15 & \textbf{29.2}$\pm$0.55 \\
& ROID & \textbf{23.6}$\pm$0.05 & 29.4$\pm$0.13 \\
\hline
\end{tabular}}
\vskip -0.2in
\end{table}

\textbf{Results for correlated (+mixed domains)} 
First, we consider a correlated setting, where samples are sorted by class. Since re-calculating BN statistics now even increases the error absolutely by 13.8\% to 95.8\% for a ResNet-50 on the long ImageNet-C sequence, we only consider transformers based on layer normalization and the same ResNet-26 with group normalization that was used in \cite{zhang2021memo}.

The results are presented in \Cref{fig:correlated} (left). Detailed results are further shown in \Cref{tab:correlated} and \Cref{tab:detailed-correlated} in the appendix. Even though SAR was proposed for a correlated setting, in this extreme case of sorted classes and multiple domain shifts, its performance often degrades below the source baseline. A similar trend can also be observed for RoTTA, which also does not show any substantial performance improvements. The only methods that can significantly outperform the source baseline are LAME and ROID. Since LAME tends to predict only a few classes, it performs well in the correlated setting, while drastically degrading the performance in previous scenarios. ROID, on the other hand, outperforms LAME on 3 out of 5 datasets, while also showing strong results in other settings. On the right of \Cref{fig:correlated}, we illustrate the performance for different degrees of correlation by varying the concentration parameter $\delta$ of a Dirichlet distribution \cite{zhu2021federated, gong2022note}. Prior correction and, consequently, ROID benefit from increasing correlation, as the entropy of the class prior decreases.

Lastly, we investigate the combination of temporally correlated data with mixed domains for IN-C and IN-D109. As shown in \Cref{tab:mixed-correlated}, ROID achieves significantly better and comparable results than existing methods, demonstrating its ability to perform in all scenarios of universal TTA.

\textbf{Results for single sample TTA}
Updating the model using a single test sample not only yields noisy gradients, but also prevents an accurate estimation of the BN statistics, resulting in a performance degradation. While \cite{dobler2023robust, GTTA} use a small buffer to store the last $b$ test samples on the device, this comes with a trade-off between efficiency and accurate BN statistics. To circumvent this issue, we propose to use networks that do not employ BN layers, such as VisionTransformer \cite{dosovitskiy2020image}. These networks allow to recover the batch TTA setting by simply accumulating the gradients of the last $b$ test samples before updating the model. As shown in \Cref{tab:appendix_efficiency_memory}, this provides the same results as before, with no computational overhead and significantly reduced memory requirements.

\subsection{Ablation studies}
In \Cref{sec:appendix_ablations}, we further analyze the efficiency, catastrophic forgetting, and the momentum $\alpha$ used for weight ensembling. We find that ROID successfully maintains its knowledge about the initial training domain while being computationally efficient.

\textbf{Component analysis} In \Cref{tab:ablation-components-settings}, we analyze the components of ROID. In general, the component analysis underscores our primary hypotheses and findings. Certainty and diversity based loss weighting helps in all scenarios by mitigating the development of a model bias. Weight ensembling demonstrates its effectiveness in settings where the model has to adapt sequentially to multiple narrow distributions, such as in the \textit{continual} and \textit{correlated} setting. It does not contribute, when a broad distribution is present (\textit{mixed domains}). For the difficult adaptation in correlated settings, weight ensembling also serves as a corrective measure. It addresses suboptimal adaptations over time by continually incorporating a small percentage of the source weights. Finally, prior correction shows its strong suits in \textit{correlated} settings and upholds performance when a uniform class distribution is present. Further details and discussions are located in \ref{sec:appendix_component_analysis}.

\begin{table}[t]
\renewcommand{\arraystretch}{1.2}
\centering
\caption{Average online classification error rate~(\%) over 5 runs for different configurations and settings.} \label{tab:ablation-components-settings}
\vskip -0.1in
\scalebox{0.8}{
\tabcolsep4pt
\begin{tabular}{l| cccc}\hline
Method  & \rotatebox[origin=c]{70}{ continual } & \rotatebox[origin=c]{70}{mixed} & \rotatebox[origin=c]{70}{ correlated } & \rotatebox[origin=c]{70}{ mix. + corr. } \\
\hline
  Source                    & 61.7 & 57.7 & 54.6 & 48.8 \\
 \hline
 SLR             & 52.6 & \textcolor{red}{66.7} & \textcolor{red}{80.0} & \textcolor{red}{88.1} \\
 + Loss weighting  & 46.1 & 46.4 & \textcolor{red}{60.4} & \textcolor{red}{61.1} \\
 + Weight ensembling            & 45.0 & 46.9 & 46.7 & 44.9 \\
 + Consistency   & \textbf{43.9} & 46.0 & 45.7 & 43.8 \\
 + Prior correction & \textbf{43.9} & \textbf{45.9} & \textbf{26.8} & \textbf{23.5} \\
\hline
\end{tabular}}
\vskip -0.2in
\end{table}


\section{Conclusion} 
In this work, we derive all practically relevant settings and denote this as \textit{universal TTA}. By further highlighting several challenges which can arise when conducting self-training during test-time, namely the loss of generalization, model bias, and trivial solutions, we introduce a new TTA method: ROID. To retain generalization, ROID continually weight-averages the source and adapted model. For promoting stability and encourage diverse predictions, a certainty and diversity weighted SLR loss is used. To compensate for prior shifts that can occur during test-time, a novel adaptive prior correction scheme is proposed. We set new standards in the field of online universal TTA.

{\small
\bibliographystyle{ieee_fullname}
\bibliography{egbib}

\begin{thebibliography}{10}\itemsep=-1pt

\bibitem{MT3}
Alexander Bartler, Andre B{\"u}hler, Felix Wiewel, Mario D{\"o}bler, and Bin
  Yang.
\newblock Mt3: Meta test-time training for self-supervised test-time adaption.
\newblock In {\em International Conference on Artificial Intelligence and
  Statistics}, pages 3080--3090. PMLR, 2022.

\bibitem{boudiaf2022parameter}
Malik Boudiaf, Romain Mueller, Ismail Ben~Ayed, and Luca Bertinetto.
\newblock Parameter-free online test-time adaptation.
\newblock In {\em Proceedings of the IEEE/CVF Conference on Computer Vision and
  Pattern Recognition}, pages 8344--8353, 2022.

\bibitem{chen2020homm}
Chao Chen, Zhihang Fu, Zhihong Chen, Sheng Jin, Zhaowei Cheng, Xinyu Jin, and
  Xian-Sheng Hua.
\newblock Homm: Higher-order moment matching for unsupervised domain
  adaptation.
\newblock In {\em Proceedings of the AAAI conference on artificial
  intelligence}, volume~34, pages 3422--3429, 2020.

\bibitem{chen2022contrastive}
Dian Chen, Dequan Wang, Trevor Darrell, and Sayna Ebrahimi.
\newblock Contrastive test-time adaptation.
\newblock In {\em Proceedings of the IEEE/CVF Conference on Computer Vision and
  Pattern Recognition}, pages 295--305, 2022.

\bibitem{dobler2023robust}
Mario D{\"o}bler, Robert~A Marsden, and Bin Yang.
\newblock Robust mean teacher for continual and gradual test-time adaptation.
\newblock In {\em Proceedings of the IEEE/CVF Conference on Computer Vision and
  Pattern Recognition}, pages 7704--7714, 2023.

\bibitem{dosovitskiy2020image}
Alexey Dosovitskiy, Lucas Beyer, Alexander Kolesnikov, Dirk Weissenborn,
  Xiaohua Zhai, Thomas Unterthiner, Mostafa Dehghani, Matthias Minderer, Georg
  Heigold, Sylvain Gelly, et~al.
\newblock An image is worth 16x16 words: Transformers for image recognition at
  scale.
\newblock {\em arXiv preprint arXiv:2010.11929}, 2020.

\bibitem{frankle2020linear}
Jonathan Frankle, Gintare~Karolina Dziugaite, Daniel Roy, and Michael Carbin.
\newblock Linear mode connectivity and the lottery ticket hypothesis.
\newblock In {\em International Conference on Machine Learning}, pages
  3259--3269. PMLR, 2020.

\bibitem{french2018selfensembling}
Geoff French, Michal Mackiewicz, and Mark Fisher.
\newblock Self-ensembling for visual domain adaptation.
\newblock In {\em International Conference on Learning Representations}, 2018.

\bibitem{ganin2016domain}
Yaroslav Ganin, Evgeniya Ustinova, Hana Ajakan, Pascal Germain, Hugo
  Larochelle, Fran{\c{c}}ois Laviolette, Mario Marchand, and Victor Lempitsky.
\newblock Domain-adversarial training of neural networks.
\newblock {\em The journal of machine learning research}, 17(1):2096--2030,
  2016.

\bibitem{gao2022back}
Jin Gao, Jialing Zhang, Xihui Liu, Trevor Darrell, Evan Shelhamer, and Dequan
  Wang.
\newblock Back to the source: Diffusion-driven test-time adaptation.
\newblock {\em arXiv preprint arXiv:2207.03442}, 2022.

\bibitem{gong2022note}
Taesik Gong, Jongheon Jeong, Taewon Kim, Yewon Kim, Jinwoo Shin, and Sung-Ju
  Lee.
\newblock Note: Robust continual test-time adaptation against temporal
  correlation.
\newblock In {\em Advances in Neural Information Processing Systems (NeurIPS)},
  2022.

\bibitem{hendrycks2021many}
Dan Hendrycks, Steven Basart, Norman Mu, Saurav Kadavath, Frank Wang, Evan
  Dorundo, Rahul Desai, Tyler Zhu, Samyak Parajuli, Mike Guo, et~al.
\newblock The many faces of robustness: A critical analysis of
  out-of-distribution generalization.
\newblock In {\em Proceedings of the IEEE/CVF International Conference on
  Computer Vision}, pages 8340--8349, 2021.

\bibitem{hendrycks2019benchmarking}
Dan Hendrycks and Thomas Dietterich.
\newblock Benchmarking neural network robustness to common corruptions and
  perturbations.
\newblock {\em arXiv preprint arXiv:1903.12261}, 2019.

\bibitem{hendrycks2019augmix}
Dan Hendrycks, Norman Mu, Ekin~D Cubuk, Barret Zoph, Justin Gilmer, and Balaji
  Lakshminarayanan.
\newblock Augmix: A simple data processing method to improve robustness and
  uncertainty.
\newblock {\em arXiv preprint arXiv:1912.02781}, 2019.

\bibitem{CYCADA}
Judy Hoffman, Eric Tzeng, Taesung Park, Jun-Yan Zhu, Phillip Isola, Kate
  Saenko, Alexei Efros, and Trevor Darrell.
\newblock Cycada: Cycle-consistent adversarial domain adaptation.
\newblock In {\em International conference on machine learning}, pages
  1989--1998. PMLR, 2018.

\bibitem{izmailov2018averaging}
Pavel Izmailov, Dmitrii Podoprikhin, Timur Garipov, Dmitry Vetrov, and
  Andrew~Gordon Wilson.
\newblock Averaging weights leads to wider optima and better generalization.
\newblock {\em arXiv preprint arXiv:1803.05407}, 2018.

\bibitem{kang2019contrastive}
Guoliang Kang, Lu Jiang, Yi Yang, and Alexander~G Hauptmann.
\newblock Contrastive adaptation network for unsupervised domain adaptation.
\newblock In {\em Proceedings of the IEEE/CVF Conference on Computer Vision and
  Pattern Recognition}, pages 4893--4902, 2019.

\bibitem{kirkpatrick2017overcoming}
James Kirkpatrick, Razvan Pascanu, Neil Rabinowitz, Joel Veness, Guillaume
  Desjardins, Andrei~A Rusu, Kieran Milan, John Quan, Tiago Ramalho, Agnieszka
  Grabska-Barwinska, et~al.
\newblock Overcoming catastrophic forgetting in neural networks.
\newblock {\em Proceedings of the national academy of sciences},
  114(13):3521--3526, 2017.

\bibitem{krizhevsky2009learning}
Alex Krizhevsky, Geoffrey Hinton, et~al.
\newblock Learning multiple layers of features from tiny images.
\newblock 2009.

\bibitem{liang2020we}
Jian Liang, Dapeng Hu, and Jiashi Feng.
\newblock Do we really need to access the source data? source hypothesis
  transfer for unsupervised domain adaptation.
\newblock In {\em International Conference on Machine Learning}, pages
  6028--6039. PMLR, 2020.

\bibitem{liu2021cycle}
Hong Liu, Jianmin Wang, and Mingsheng Long.
\newblock Cycle self-training for domain adaptation.
\newblock {\em Advances in Neural Information Processing Systems},
  34:22968--22981, 2021.

\bibitem{liu2021ttt++}
Yuejiang Liu, Parth Kothari, Bastien van Delft, Baptiste Bellot-Gurlet, Taylor
  Mordan, and Alexandre Alahi.
\newblock Ttt++: When does self-supervised test-time training fail or thrive?
\newblock {\em Advances in Neural Information Processing Systems}, 34, 2021.

\bibitem{liu2021swin}
Ze Liu, Yutong Lin, Yue Cao, Han Hu, Yixuan Wei, Zheng Zhang, Stephen Lin, and
  Baining Guo.
\newblock Swin transformer: Hierarchical vision transformer using shifted
  windows.
\newblock In {\em Proceedings of the IEEE/CVF International Conference on
  Computer Vision}, pages 10012--10022, 2021.

\bibitem{marsden2022contrastive}
Robert~A Marsden, Alexander Bartler, Mario D{\"o}bler, and Bin Yang.
\newblock Contrastive learning and self-training for unsupervised domain
  adaptation in semantic segmentation.
\newblock In {\em 2022 International Joint Conference on Neural Networks
  (IJCNN)}, pages 1--8. IEEE, 2022.

\bibitem{GTTA}
Robert~A Marsden, Mario D{\"o}bler, and Bin Yang.
\newblock Introducing intermediate domains for effective self-training during
  test-time.
\newblock {\em arXiv preprint arXiv:2208.07736}, 2022.

\bibitem{marsden2022continual}
Robert~A Marsden, Felix Wiewel, Mario D{\"o}bler, Yang Yang, and Bin Yang.
\newblock Continual unsupervised domain adaptation for semantic segmentation
  using a class-specific transfer.
\newblock In {\em 2022 International Joint Conference on Neural Networks
  (IJCNN)}, pages 1--8. IEEE, 2022.

\bibitem{mccloskey1989catastrophic}
Michael McCloskey and Neal~J Cohen.
\newblock Catastrophic interference in connectionist networks: The sequential
  learning problem.
\newblock In {\em Psychology of learning and motivation}, volume~24, pages
  109--165. Elsevier, 1989.

\bibitem{IAST}
Ke Mei, Chuang Zhu, Jiaqi Zou, and Shanghang Zhang.
\newblock Instance adaptive self-training for unsupervised domain adaptation.
\newblock {\em arXiv preprint arXiv:2008.12197}, 2020.

\bibitem{mintun2021interaction}
Eric Mintun, Alexander Kirillov, and Saining Xie.
\newblock On interaction between augmentations and corruptions in natural
  corruption robustness.
\newblock {\em Advances in Neural Information Processing Systems}, 34, 2021.

\bibitem{mirza2022norm}
M~Jehanzeb Mirza, Jakub Micorek, Horst Possegger, and Horst Bischof.
\newblock The norm must go on: Dynamic unsupervised domain adaptation by
  normalization.
\newblock In {\em Proceedings of the IEEE/CVF Conference on Computer Vision and
  Pattern Recognition}, pages 14765--14775, 2022.

\bibitem{muandet2013domain}
Krikamol Muandet, David Balduzzi, and Bernhard Sch{\"o}lkopf.
\newblock Domain generalization via invariant feature representation.
\newblock In {\em International Conference on Machine Learning}, pages 10--18.
  PMLR, 2013.

\bibitem{mummadi2021test}
Chaithanya~Kumar Mummadi, Robin Hutmacher, Kilian Rambach, Evgeny Levinkov,
  Thomas Brox, and Jan~Hendrik Metzen.
\newblock Test-time adaptation to distribution shift by confidence maximization
  and input transformation.
\newblock {\em arXiv preprint arXiv:2106.14999}, 2021.

\bibitem{niu2022efficient}
Shuaicheng Niu, Jiaxiang Wu, Yifan Zhang, Yaofo Chen, Shijian Zheng, Peilin
  Zhao, and Mingkui Tan.
\newblock Efficient test-time model adaptation without forgetting.
\newblock In {\em International conference on machine learning}, pages
  16888--16905. PMLR, 2022.

\bibitem{niu2023towards}
Shuaicheng Niu, Jiaxiang Wu, Yifan Zhang, Zhiquan Wen, Yaofo Chen, Peilin Zhao,
  and Mingkui Tan.
\newblock Towards stable test-time adaptation in dynamic wild world.
\newblock In {\em The Eleventh International Conference on Learning
  Representations}, 2023.

\bibitem{peng2019moment}
Xingchao Peng, Qinxun Bai, Xide Xia, Zijun Huang, Kate Saenko, and Bo Wang.
\newblock Moment matching for multi-source domain adaptation.
\newblock In {\em Proceedings of the IEEE/CVF international conference on
  computer vision}, pages 1406--1415, 2019.

\bibitem{quinonero2008dataset}
Joaquin Qui{\~n}onero-Candela, Masashi Sugiyama, Anton Schwaighofer, and Neil~D
  Lawrence.
\newblock {\em Dataset shift in machine learning}.
\newblock Mit Press, 2008.

\bibitem{radford2021learning}
Alec Radford, Jong~Wook Kim, Chris Hallacy, Aditya Ramesh, Gabriel Goh,
  Sandhini Agarwal, Girish Sastry, Amanda Askell, Pamela Mishkin, Jack Clark,
  et~al.
\newblock Learning transferable visual models from natural language
  supervision.
\newblock In {\em International Conference on Machine Learning}, pages
  8748--8763. PMLR, 2021.

\bibitem{royer2015classifier}
Amelie Royer and Christoph~H Lampert.
\newblock Classifier adaptation at prediction time.
\newblock In {\em Proceedings of the IEEE Conference on Computer Vision and
  Pattern Recognition}, pages 1401--1409, 2015.

\bibitem{rusak2022imagenet}
Evgenia Rusak, Steffen Schneider, Peter~Vincent Gehler, Oliver Bringmann,
  Wieland Brendel, and Matthias Bethge.
\newblock Imagenet-d: A new challenging robustness dataset inspired by domain
  adaptation.
\newblock In {\em ICML 2022 Shift Happens Workshop}, 2022.

\bibitem{saito2019semi}
Kuniaki Saito, Donghyun Kim, Stan Sclaroff, Trevor Darrell, and Kate Saenko.
\newblock Semi-supervised domain adaptation via minimax entropy.
\newblock In {\em Proceedings of the IEEE/CVF International Conference on
  Computer Vision}, pages 8050--8058, 2019.

\bibitem{sankaranarayanan2018generate}
Swami Sankaranarayanan, Yogesh Balaji, Carlos~D Castillo, and Rama Chellappa.
\newblock Generate to adapt: Aligning domains using generative adversarial
  networks.
\newblock In {\em Proceedings of the IEEE conference on computer vision and
  pattern recognition}, pages 8503--8512, 2018.

\bibitem{schneider2020improving}
Steffen Schneider, Evgenia Rusak, Luisa Eck, Oliver Bringmann, Wieland Brendel,
  and Matthias Bethge.
\newblock Improving robustness against common corruptions by covariate shift
  adaptation.
\newblock {\em Advances in Neural Information Processing Systems},
  33:11539--11551, 2020.

\bibitem{sun2016deep}
Baochen Sun and Kate Saenko.
\newblock Deep coral: Correlation alignment for deep domain adaptation.
\newblock In {\em European conference on computer vision}, pages 443--450.
  Springer, 2016.

\bibitem{TTT}
Yu Sun, Xiaolong Wang, Zhuang Liu, John Miller, Alexei Efros, and Moritz Hardt.
\newblock Test-time training with self-supervision for generalization under
  distribution shifts.
\newblock In {\em International Conference on Machine Learning}, pages
  9229--9248. PMLR, 2020.

\bibitem{tarvainen2017mean}
Antti Tarvainen and Harri Valpola.
\newblock Mean teachers are better role models: Weight-averaged consistency
  targets improve semi-supervised deep learning results.
\newblock {\em Advances in neural information processing systems}, 30, 2017.

\bibitem{tobin2017domain}
Josh Tobin, Rachel Fong, Alex Ray, Jonas Schneider, Wojciech Zaremba, and
  Pieter Abbeel.
\newblock Domain randomization for transferring deep neural networks from
  simulation to the real world.
\newblock In {\em 2017 IEEE/RSJ international conference on intelligent robots
  and systems (IROS)}, pages 23--30. IEEE, 2017.

\bibitem{tranheden2021dacs}
Wilhelm Tranheden, Viktor Olsson, Juliano Pinto, and Lennart Svensson.
\newblock Dacs: Domain adaptation via cross-domain mixed sampling.
\newblock In {\em Proceedings of the IEEE/CVF Winter Conference on Applications
  of Computer Vision}, pages 1379--1389, 2021.

\bibitem{tremblay2018training}
Jonathan Tremblay, Aayush Prakash, David Acuna, Mark Brophy, Varun Jampani, Cem
  Anil, Thang To, Eric Cameracci, Shaad Boochoon, and Stan Birchfield.
\newblock Training deep networks with synthetic data: Bridging the reality gap
  by domain randomization.
\newblock In {\em Proceedings of the IEEE conference on computer vision and
  pattern recognition workshops}, pages 969--977, 2018.

\bibitem{tzeng2017adversarial}
Eric Tzeng, Judy Hoffman, Kate Saenko, and Trevor Darrell.
\newblock Adversarial discriminative domain adaptation.
\newblock In {\em Proceedings of the IEEE conference on computer vision and
  pattern recognition}, pages 7167--7176, 2017.

\bibitem{ADVENT}
Tuan-Hung Vu, Himalaya Jain, Maxime Bucher, Matthieu Cord, and Patrick
  P{\'e}rez.
\newblock Advent: Adversarial entropy minimization for domain adaptation in
  semantic segmentation.
\newblock In {\em Proceedings of the IEEE conference on computer vision and
  pattern recognition}, pages 2517--2526, 2019.

\bibitem{wang2021tent}
Dequan Wang, Evan Shelhamer, Shaoteng Liu, Bruno Olshausen, and Trevor Darrell.
\newblock Tent: Fully test-time adaptation by entropy minimization.
\newblock In {\em International Conference on Learning Representations}, 2021.

\bibitem{wang2019learning}
Haohan Wang, Songwei Ge, Zachary Lipton, and Eric~P Xing.
\newblock Learning robust global representations by penalizing local predictive
  power.
\newblock {\em Advances in Neural Information Processing Systems}, 32, 2019.

\bibitem{wang2022continual}
Qin Wang, Olga Fink, Luc Van~Gool, and Dengxin Dai.
\newblock Continual test-time domain adaptation.
\newblock In {\em Proceedings of the IEEE/CVF Conference on Computer Vision and
  Pattern Recognition}, pages 7201--7211, 2022.

\bibitem{wang2019symmetric}
Yisen Wang, Xingjun Ma, Zaiyi Chen, Yuan Luo, Jinfeng Yi, and James Bailey.
\newblock Symmetric cross entropy for robust learning with noisy labels.
\newblock In {\em Proceedings of the IEEE/CVF International Conference on
  Computer Vision}, pages 322--330, 2019.

\bibitem{wilson2020survey}
Garrett Wilson and Diane~J Cook.
\newblock A survey of unsupervised deep domain adaptation.
\newblock {\em ACM Transactions on Intelligent Systems and Technology (TIST)},
  11(5):1--46, 2020.

\bibitem{wortsman2022model}
Mitchell Wortsman, Gabriel Ilharco, Samir~Ya Gadre, Rebecca Roelofs, Raphael
  Gontijo-Lopes, Ari~S Morcos, Hongseok Namkoong, Ali Farhadi, Yair Carmon,
  Simon Kornblith, et~al.
\newblock Model soups: averaging weights of multiple fine-tuned models improves
  accuracy without increasing inference time.
\newblock In {\em International Conference on Machine Learning}, pages
  23965--23998. PMLR, 2022.

\bibitem{ACE}
Zuxuan Wu, Xin Wang, Joseph~E Gonzalez, Tom Goldstein, and Larry~S Davis.
\newblock Ace: adapting to changing environments for semantic segmentation.
\newblock In {\em Proceedings of the IEEE International Conference on Computer
  Vision}, pages 2121--2130, 2019.

\bibitem{xie2017aggregated}
Saining Xie, Ross Girshick, Piotr Doll{\'a}r, Zhuowen Tu, and Kaiming He.
\newblock Aggregated residual transformations for deep neural networks.
\newblock In {\em Proceedings of the IEEE conference on computer vision and
  pattern recognition}, pages 1492--1500, 2017.

\bibitem{yan2017mind}
Hongliang Yan, Yukang Ding, Peihua Li, Qilong Wang, Yong Xu, and Wangmeng Zuo.
\newblock Mind the class weight bias: Weighted maximum mean discrepancy for
  unsupervised domain adaptation.
\newblock In {\em Proceedings of the IEEE conference on computer vision and
  pattern recognition}, pages 2272--2281, 2017.

\bibitem{yuan2023robust}
Longhui Yuan, Binhui Xie, and Shuang Li.
\newblock Robust test-time adaptation in dynamic scenarios.
\newblock In {\em Proceedings of the IEEE/CVF Conference on Computer Vision and
  Pattern Recognition}, pages 15922--15932, 2023.

\bibitem{zagoruyko2016wide}
Sergey Zagoruyko and Nikos Komodakis.
\newblock Wide residual networks.
\newblock {\em arXiv preprint arXiv:1605.07146}, 2016.

\bibitem{zhang2021memo}
Marvin Zhang, Sergey Levine, and Chelsea Finn.
\newblock Memo: Test time robustness via adaptation and augmentation.
\newblock {\em arXiv preprint arXiv:2110.09506}, 2021.

\bibitem{zhu2021federated}
Hangyu Zhu, Jinjin Xu, Shiqing Liu, and Yaochu Jin.
\newblock Federated learning on non-iid data: A survey.
\newblock {\em Neurocomputing}, 465:371--390, 2021.

\bibitem{CRST}
Yang Zou, Zhiding Yu, Xiaofeng Liu, BVK Kumar, and Jinsong Wang.
\newblock Confidence regularized self-training.
\newblock In {\em Proceedings of the IEEE International Conference on Computer
  Vision}, pages 5982--5991, 2019.

\end{thebibliography}
}

\newpage
\appendix
\onecolumn
\section{Additional analysis}
\subsection{Loss of generalization}
\label{sec:appendix_generalization}
\begin{figure}[h!]
    \centering
    \includegraphics[width=0.85\textwidth]{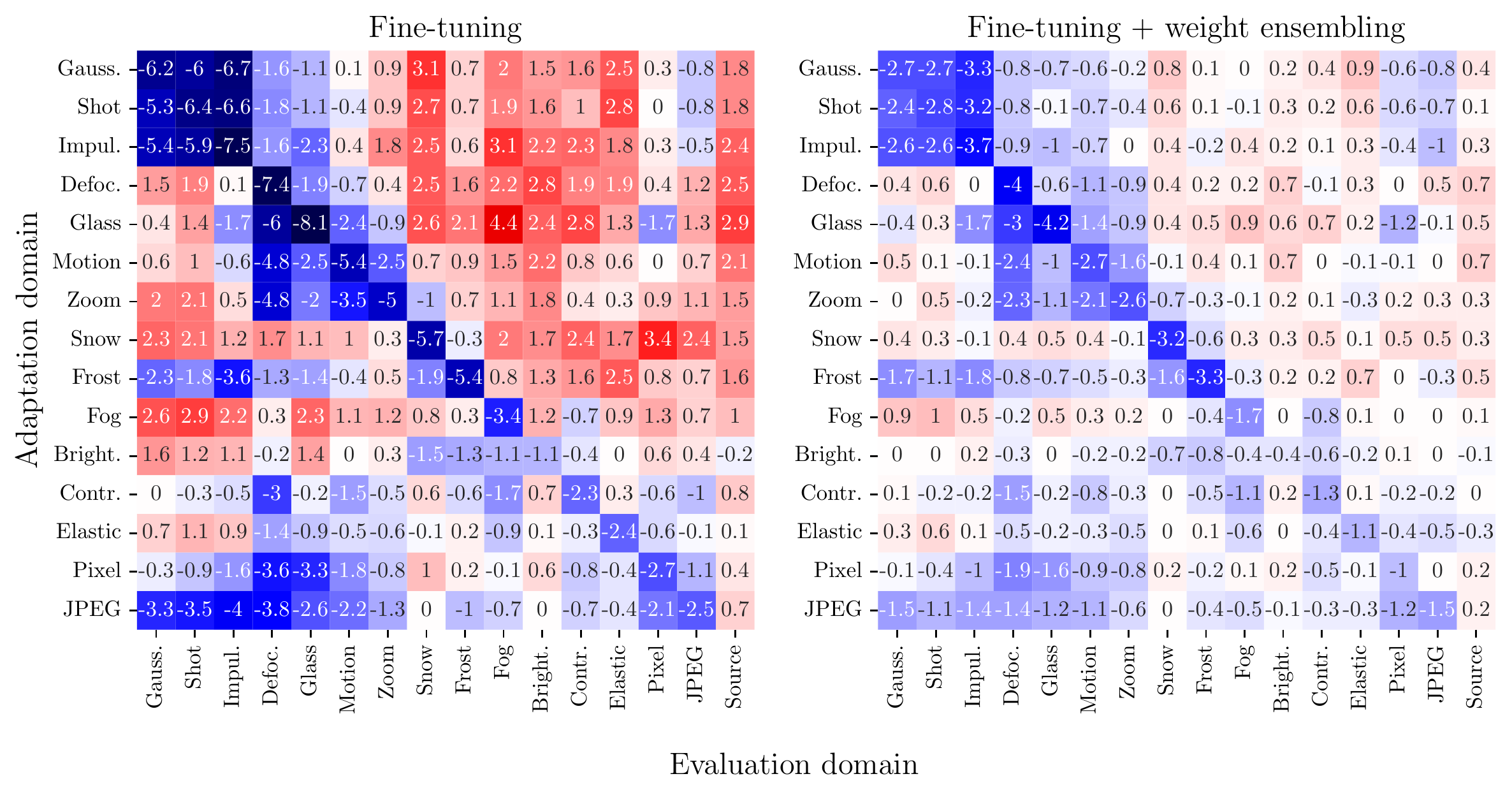}
    \vskip -0.2in
    \includegraphics[width=0.85\textwidth]{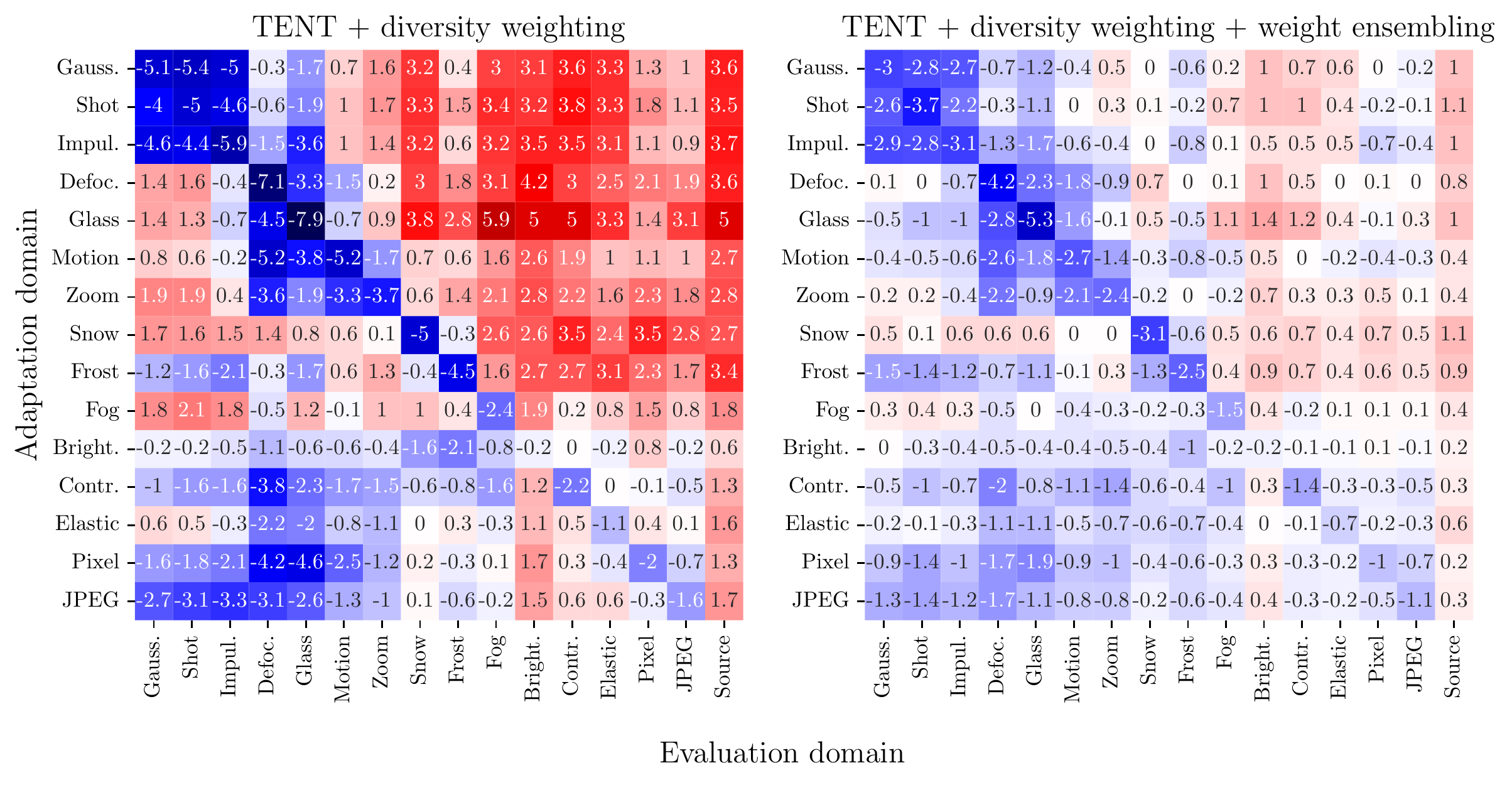}
    \vskip -0.05in
    \caption{Difference of error for a moderate and a stronger model adaptation corresponding to a learning rate of $10^{-4}$ and $10^{-3}$, respectively. The first row examines supervised fine-tuning, while the second row considers diversity-regularized self-training. The right column further illustrates the effect of adding our weight ensembling approach. All experiments are conducted using an ImageNet pre-trained ResNet-50 that is adapted using 40,000 samples of one of the corruptions from ImageNet-C. The model is then evaluated on the remaining 10,000 samples for all corruptions as well as the source domain. Adapting the model on a potentially narrow distribution can clearly degrade its generalization capabilities. Adding weight ensembling helps to mitigate the loss of generalization as well as catastrophic forgetting.}
    \label{fig:generalization_appendix}
\end{figure}
Since adapting a model to a target domain effectively means moving the model from its initial source parameterization to a parameterization that better models the current target distribution, this should trigger a loss of generalization when the target distribution is narrow. While we have already shown in \Cref{sec:self_training_analysis} that a generalization loss occurs when performing self-training in the form of entropy minimization, this should also hold when our certainty and diversity weighting from \Cref{sec:weighting} is further added, or when fine-tuning the model in a supervised manner. 

To demonstrate the previous points, we adopt the same setup as before, i.e., we use an ImageNet pre-trained ResNet-50 and adapt the model with 40,000 samples of one of the corruptions from ImageNet-C. Afterwards, the model is evaluated for each corruption and the source domain on the remaining 10,000 samples. \Cref{fig:generalization_appendix} illustrates the difference of error for a moderate and a stronger adapted model, corresponding to a learning rate of $10^{-4}$ and $10^{-3}$, respectively. Depending on the investigated corruption, not only fine-tuning but also diversity-regularized self-training result in an increased error on other corruptions, indicating a loss of generalization. This demonstrates the risks of model adaptation in a potentially unknown environment. Using our proposed weight ensembling, a loss of generalization and catastrophic forgetting can mostly be mitigated.

\subsection{Model bias and trivial solutions}
\label{sec:appendix_error_accumulation}
As stated in \Cref{sec:self_training_analysis}, a critical factor for successful TTA is stability. Current methods for online TTA mostly leverage self-training to adapt the model to the current domain shift, showing great performance on short test sequences \cite{wang2021tent, wang2022continual, dobler2023robust, niu2023towards}. However, if self-training is utilized without any proper regularization, the model is likely to become biased after a while. In the worst case, the bias can even evolve into a trivial solution, where the model only predicts a small subset of classes. In this section, we first demonstrate the aforementioned points for TENT, which exploits entropy minimization for model adaptation. Then, we investigate the behaviour of current state-of-the-art methods, revealing some inefficiencies to effectively counter model bias during test-time. 

\begin{figure}[H]
    \centering
    \includegraphics[scale=0.66]{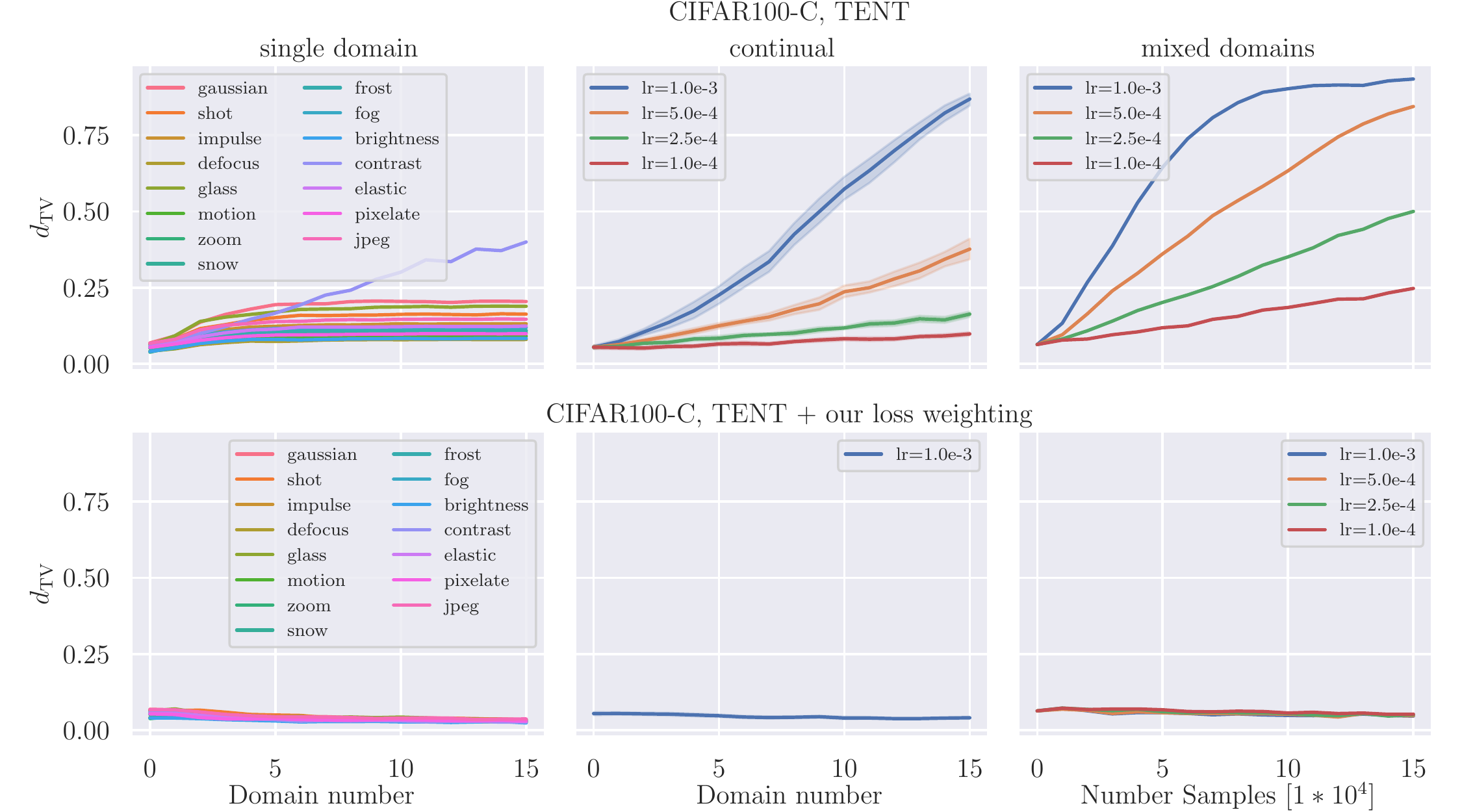}
    \vskip 0.3in
    \includegraphics[scale=0.66]{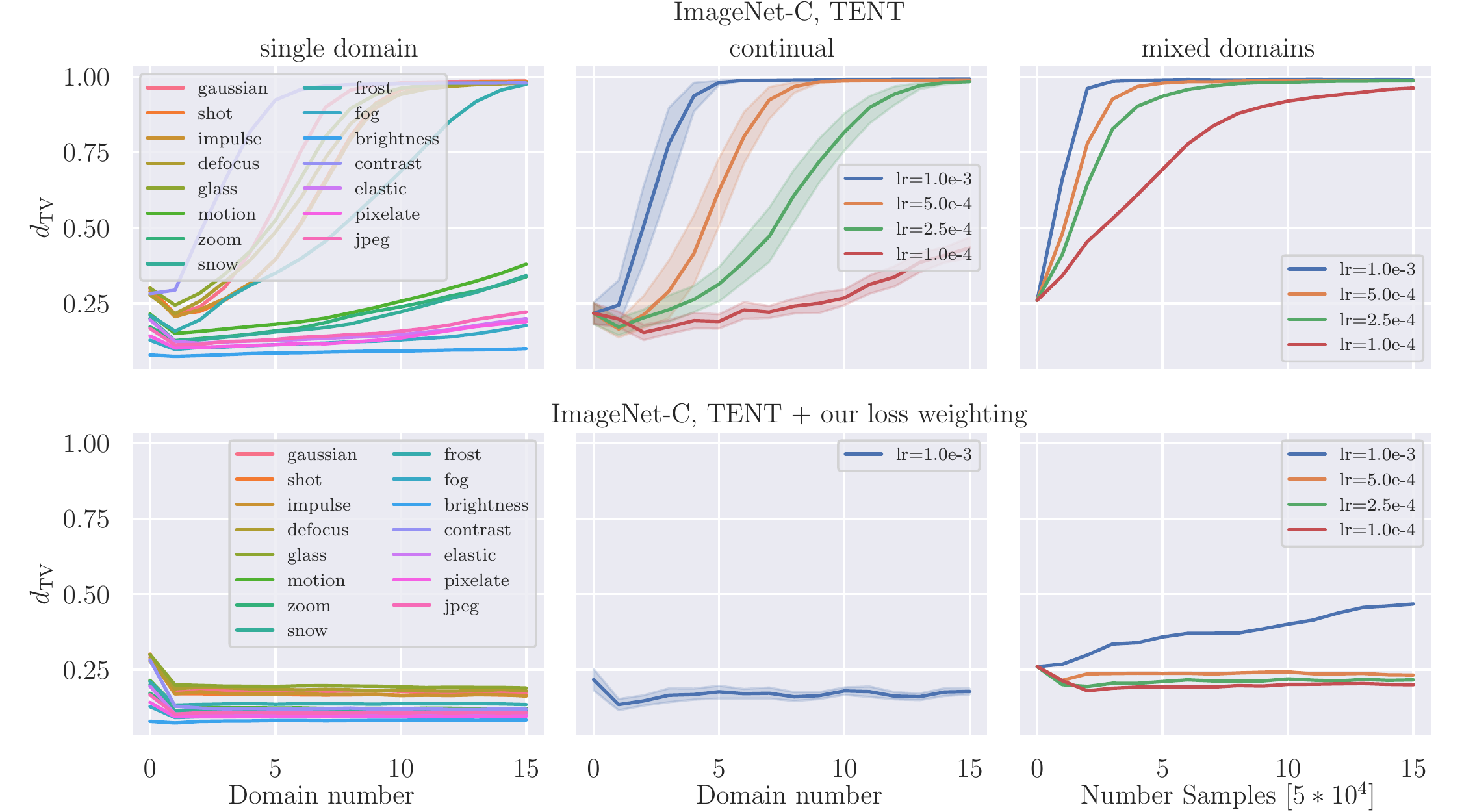}
    \caption{Illustration of the total variation distance of TENT on CIFAR100-C and ImageNet-C at severity level 5 without (first row) and with (second row) our loss weighting. The model is adapted to a single domain (left), in the continual setting (middle) using 15 randomly ordered domain sequences, and the mixed domains setting (right). Unless otherwise stated, TENT's default learning rates of $1.0\mathrm{e}^{-3}$ and $2.5\mathrm{e}^{-4}$ are used. Comparing the left and middle column of CIFAR100-C, it becomes obvious that domain shifts can promote the occurrence of trivial solutions. In case of mixed domains, model bias and trivial solutions occur even faster for both datasets. In contrast, using TENT with our loss weighting prevents the model from becoming biased in almost all settings.}
    \label{fig:tvd}
\end{figure}

\paragraph{Long test sequences promote model bias and domain shifts can trigger trivial solutions}
To investigate whether the model is becoming biased or degrades to a trivial solution during the adaptation, we consider the total variation distance (TVD). It measures the deviation between the actual class prior and the predicted prior. The TVD is defined as 
\begin{equation}
d_{\mathrm{TV}}(\bm{p}, \hat{\bm{p}}) = \frac{1}{2} \sum_{i=1}^C |p_i - \hat{p}_i|,    
\end{equation}
where $p_i$ and $\hat{p}_i$ are the true and predicted prior probability for class $i$, respectively. If the TVD is calculated along the test sequence, it can also indirectly show the occurrence of error accumulation, since it is a lower bound of the error of the pseudo-labels \cite{liu2021cycle}. Since TENT reports good results for adapting a model to a single domain, we begin our analysis with the same setting and only vary the length of the test sequence by repeating each domain several times. Specifically, we use ImageNet-C with 50,000 samples per corruption and CIFAR100-C with 10,000 samples per corruption (both at severity level 5). Following TENT, we utilize a ResNet-50 with a learning rate $\mathrm{lr}=2.5\mathrm{e}^{-4}$ for ImageNet-C and a ResNeXt-29 with $\mathrm{lr}=0.001$ for CIFAR100-C. As shown on the left side of \Cref{fig:tvd}, TENT quickly deteriorates to a trivial solution for half of the corruptions of ImageNet-C, while developing a growing bias for the other half. In case of CIFAR100-C, TENT initially deteriorates slightly but then remains stable for most of the corruptions. To study the impact of multiple domain shifts, which is a quite common setting in practice, we leverage all 15 corruption types and create 15 randomly ordered domain sequences. The results for this setting, including different learning rates, are depicted in the middle of \Cref{fig:tvd}. Since the TVD now steadily increases in all settings, it becomes clear that domain shifts can explicitly enhance model bias and lead to trivial solutions. If the \textit{domain non-stationarity} is further increased to its maximum, where consecutive test samples are likely to originate from different domains, the TVD increases even more rapidly (right side of \Cref{fig:tvd}). Now, by equipping TENT with our certainty and diversity based loss weighting, stable adaptation across all previously considered settings and a wider range of learning rates is possible. The only exception to this is ImageNet-C in the \textit{mixed domains} TTA setting with a learning rate four times higher than the default. This clearly demonstrates that maintaining diversity is crucial in TTA.

\begin{figure}[H]
    \centering
    \includegraphics[scale=0.49]{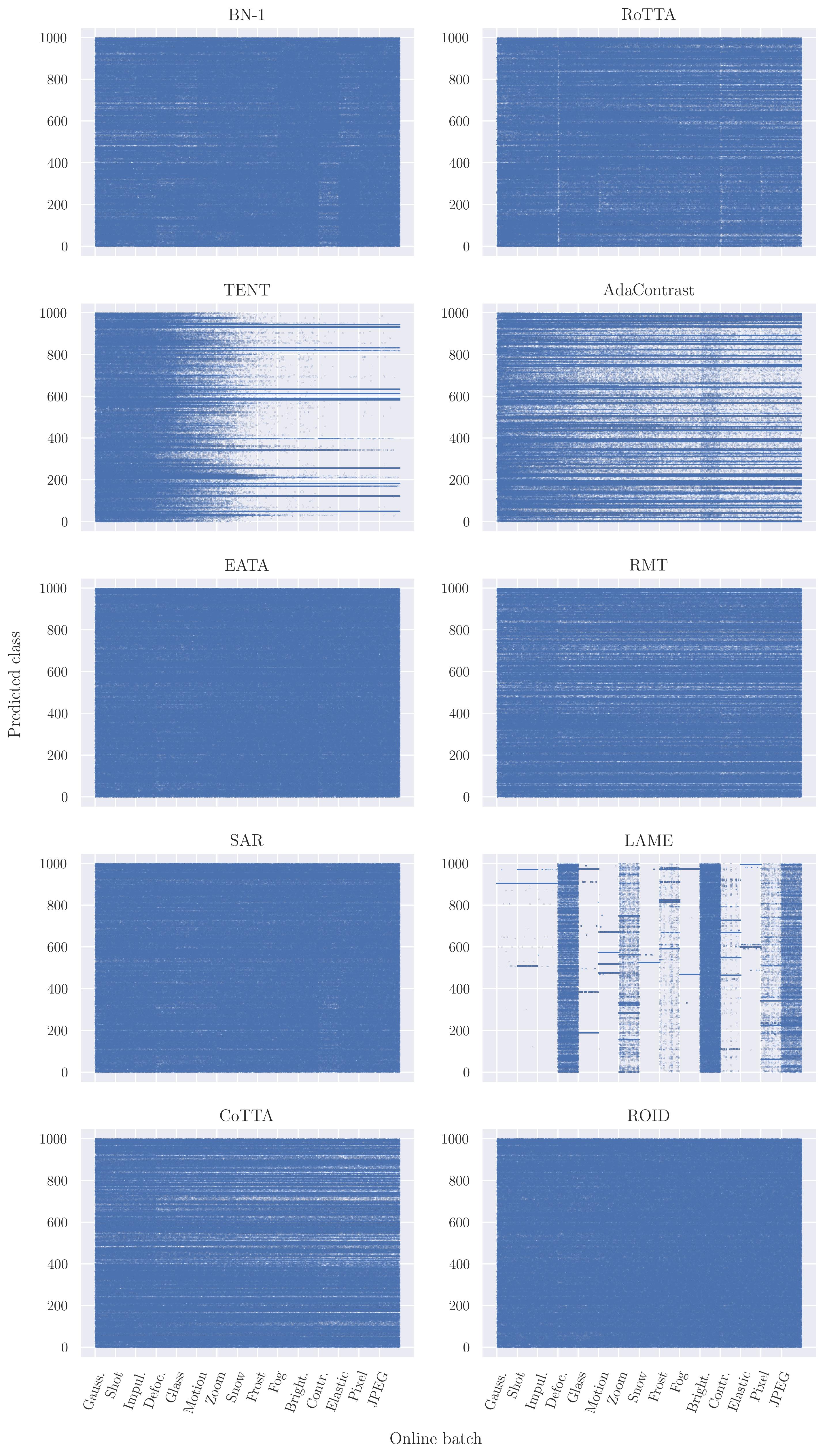}
    \caption{Illustration of the batch-wise predictions in the \textit{continual} TTA setting using a ResNet-50 and ImageNet-C with 50,000 samples per corruption.}
    \label{fig:diversity_pred}
\end{figure}

\begin{figure}[H]
    \centering
    \includegraphics[scale=0.5]{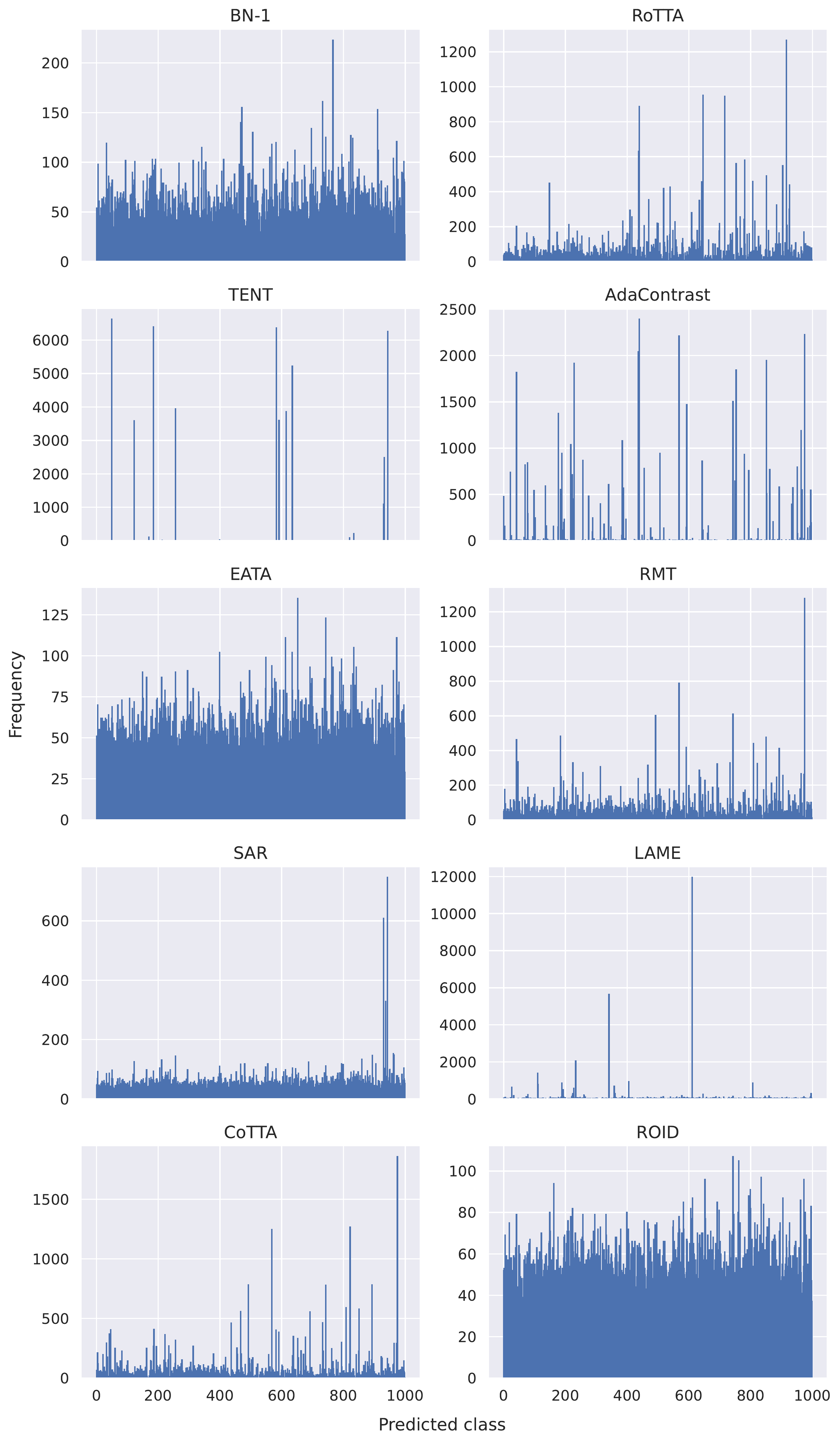}
    \caption{Frequency of the ResNet-50's predictions of the last corruption (JPEG) over the continual TTA sequence using 50,000 samples per corruption.}
    \label{fig:diversity_prior}
\end{figure}

\paragraph{Many state-of-the-art methods lack diversity}
In Figures \ref{fig:diversity_pred} and \ref{fig:diversity_prior}, we investigate existing TTA methods and our proposed method, namely ROID, in terms of diversity on the continual ImageNet-C benchmark with 50,000 samples per corruption. \Cref{fig:diversity_pred} provides a visual representation of online batch predictions across the entire continual sequence, illustrating the impact on diversity over time and the influence by different domain shifts. \Cref{fig:diversity_prior} depicts the histogram over the predicted classes for the last corruption (JPEG) after adapting the model on the complete continual sequence.

Beginning with BN--1, we observe variations in the degree of model bias induced by different domain shifts. Corruptions where the performance of BN--1 is relatively bad, tend to show a higher model bias. Looking at TENT, a collapse can be seen after a few corruptions, resulting in predicting only a small subset of the 1,000 classes. AdaContrast also strongly lacks diversity after few corruptions. Since LAME solely corrects the model output without updating the model's parameters, the diversity of its predictions heavily relies on the specific type of domain shift. Although LAME maintains diversity for certain corruptions, such as brightness, it collapses for the majority. RoTTA shows the behavior whereby diversity temporarily diminishes for specific domain shifts, such as the transition from \textit{impulse noise} to \textit{defocus blur} and \textit{brightness} to \textit{contrast}. This behavior can likely be attributed to its robust batch normalization, which incorporates past statistics, resulting in bad statistics when past statistics differ from current ones. While SAR demonstrates better diversity than CoTTA and RMT, it still manifests a deficiency in diversity, evident, for example, in the predictions for the final corruption, where a strong bias towards a few classes exists. On the other hand, EATA and ROID with their diversity weighting effectively preserve diversity throughout the adaptation process.


\section{Ablation studies}
\label{sec:appendix_ablations}

\subsection{Architectures}
\label{sec:appendix_architectures}
\begin{table*}[!h]
\renewcommand{\arraystretch}{1.2}
\centering
\caption{Online classification error rate~(\%) for the ImageNet benchmarks in the \textit{continual} TTA setting. Common architectures and their variations are considered.} \label{tab:architectures-regular}
\vskip -0.1in
\scalebox{0.9}{
\tabcolsep4pt
\begin{tabular}{l|l|cc|cccc|cc|cc|ccc| cc}\hline
&  & \multicolumn{2}{c|}{Inception} & \multicolumn{4}{c|}{ResNet} & \multicolumn{2}{c|}{ResNeXt} & \multicolumn{2}{c|}{WideResNet} & \multicolumn{3}{c|}{DenseNet} & \multicolumn{2}{c}{RegNetY} \\ \hline
&  & v1 & v3 & 18 & 50 & 101 & 152 & 50-32x4d & 101-32x8d & 50 & 101 & 121 & 169 & 201 & 8gf & 32gf  \\
\hline
& GFLOPs  &   2 &  6 & 1.8 & 3.8 & 7.6 & 11.3 & 4.2 & 8.0 & -   &  23 & 5.7 & 6.8 & 8.6 & 8.0 & 32.3 \\
& MParams & 6.6 & 27 &  12 &  25 &  44 &   60 &  25 &  44 &  69 & 127 & 8.0 &  14 &  20 &  39 &  145 \\
\hline
\rotatebox[origin=c]{90}{IN} & Source & 30.2 & 22.7 & 30.2 & 23.9 & 22.6 & 21.7 & 22.4 & 20.7 & 21.5 & 21.2 & 25.6 & 24.4 & 23.1 & 20.0 & 19.1\\
\hline
\multirow{3}{*}{\rotatebox[origin=c]{90}{IN-C}}
& Source & 81.7 & 76.5 & 85.3 & 82.0 & 77.4 & 77.6 & 78.9 & 75.2 & 78.9 & 75.3 & 78.6 & 75.7 & 75.5 & 78.4 & 75.9  \\
& BN--1  & 70.3 & 69.6 & 72.7 & 68.6 & 66.3 & 65.9 & 67.1 & 64.1 & 66.0 & 65.6 & 68.2 & 63.7 & 63.5 & 67.7 & 64.4 \\
& ROID   & 67.8 & 64.2 & 62.2 & 54.5 & 50.4 & 49.2 & 50.9 & 46.6 & 50.7 & 49.0 & 55.8 & 51.2 & 50.6 & 50.9 & 46.4 \\
\hline
\multirow{3}{*}{\rotatebox[origin=c]{90}{IN-R}}
& Source & 63.8 & 62.2 & 67.0 & 63.8 & 60.7 & 58.7 & 62.3 & 57.4 & 61.4 & 59.6 & 62.8 & 60.4 & 59.2 & 60.0 & 57.9 \\
& BN--1  & 61.5 & \textcolor{red}{63.9} & 65.1 & 60.3 & 57.7 & 56.1 & 59.3 & 56.2 & 59.1 & 58.3 & 59.8 & 57.0 & 57.3 & 59.5 & 57.0 \\
& ROID   & 59.9 & 59.9 & 59.6 & 51.2 & 46.4 & 43.9 & 48.3 & 42.0 & 46.9 & 44.9 & 50.9 & 47.7 & 46.7 & 49.2 & 42.9 \\
\hline
\multirow{3}{*}{\rotatebox[origin=c]{90}{IN-Sk.}}
& Source & 76.9 & 73.4 & 79.8 & 75.9 & 73.0 & 71.5 & 74.5 & 70.6 & 74.7 & 71.9 & 75.8 & 72.7 & 72.3 & 73.2 & 71.6 \\
& BN--1  & 74.6 & \textcolor{red}{75.0} & 77.8 & 73.6 & 72.3 & 70.9 & 73.4 & 69.2 & \textcolor{red}{75.3} & \textcolor{red}{74.7} & 75.1 & 71.9 & 72.1 & \textcolor{red}{74.8} & 69.3 \\
& ROID   & 73.3 & 71.1 & 71.5 & 64.0 & 61.2 & 59.2 & 62.1 & 57.3 & 61.9 & 60.6 & 66.0 & 62.2 & 61.4 & 62.3 & 56.6 \\
\hline
\multirow{3}{*}{\rotatebox[origin=c]{90}{IN-D109}}
& Source & 60.7 & 58.5 & 61.8 & 58.8 & 56.1 & 55.1 & 57.4 & 54.1 & 57.2 & 55.3 & 58.3 & 56.2 & 55.5 & 55.4 & 53.7 \\
& BN--1  & 58.0 & \textcolor{red}{60.5} & 59.4 & 55.1 & 53.7 & 52.4 & 54.7 & 51.7 & 56.2 & \textcolor{red}{55.6} & 56.0 & 53.7 & 54.2 & \textcolor{red}{55.6} & 52.6 \\
& ROID   & 56.5 & 57.4 & 54.6 & 47.9 & 46.1 & 44.0 & 46.3 & 43.6 & 46.5 & 45.0 & 48.5 & 46.5 & 46.1 & 47.3 & 43.2 \\
\hline
\end{tabular}
}
\end{table*}
\begin{table*}[!h]
\renewcommand{\arraystretch}{1.2}
\centering
\caption{Online classification error rate~(\%) for the ImageNet benchmarks in the \textit{continual} TTA setting. Mobile and transformer architectures and their variations are considered (\textbf{t}iny, \textbf{s}mall, \textbf{b}ase).} \label{tab:architectures-mobile}
\vskip -0.1in
\scalebox{0.9}{
\tabcolsep4pt
\begin{tabular}{l|l|ccc|cc|cc||ccc|ccc|cc|c}\hline
&  & \multicolumn{3}{c|}{MobileNet} & \multicolumn{2}{c|}{RegNetX} & \multicolumn{2}{c||}{RegNetY} & \multicolumn{3}{c|}{Swin} & \multicolumn{3}{c|}{Swin v2} & \multicolumn{2}{c|}{ViT} & MaxViT \\ \hline
&  & v2 & v3-s & v3-l & 400mf & 800mf & 400mf & 800mf & t & s & b & t & s & b & b-16 & b-32 & t \\
\hline
& GFLOPs  & 0.30 & 0.06 & 0.22 & 0.40 & 0.80 & 0.40 & 0.80 & 4.5 & 8.7 & 15.4 & 5.9 & 11.5 & 20.3 & 16.9 & - & 5.6\\
& MParams &  3.4 &  2.5 &  5.4 &  5.2 &  7.3 &  4.3 &  6.3 &  29 &  50 &   88 &  28 &   50 &   88 &  86 &  88 &  31 \\
\hline
\rotatebox[origin=c]{90}{IN} & Source & 28.1 & 32.3 & 26.0 & 27.2 & 24.8 & 26.0 & 23.6 & 18.5 & 16.8 & 16.4 & 17.9 & 16.3 & 15.9 & 18.9 & 24.1 & 16.3 \\
\hline
\multirow{3}{*}{\rotatebox[origin=c]{90}{IN-C}}
& Source & 86.7 & 83.5 & 82.5 & 84.5 & 84.0 & 83.3 & 80.6 & 70.5 & 63.7 & 64.0 & 71.7 & 65.2 & 64.2 & 60.2 & 61.6 & 54.9 \\
& BN--1  & 77.2 & 74.7 & 73.0 & 73.7 & 72.4 & 73.2 & 70.0 &   -  &   -  &   -  &   -  &   -  &   -  &   -  &   -  & 53.4 \\
& ROID   & 66.0 & 67.7 & 64.2 & 63.8 & 61.6 & 63.9 & 59.6 & 52.9 & 48.8 & 46.8 & 54.8 & 47.8 & 47.5 & 44.9 & 52.0 & 40.0 \\
\hline
\multirow{3}{*}{\rotatebox[origin=c]{90}{IN-R}}
& Source & 69.0 & 70.7 & 65.4 & 66.4 & 65.9 & 67.0 & 64.5 & 58.7 & 55.3 & 54.3 & 60.0 & 55.9 & 54.8 & 56.0 & 58.2 & 50.6 \\
& BN--1  & 67.8 & \textcolor{red}{71.7} & \textcolor{red}{66.5} & 65.9 & 64.4 & \textcolor{red}{67.1} & 63.9 &   -  &   -  &   -  &   -  &   -  &   -  &   -  &   -  & 49.0 \\
& ROID   & 62.0 & 69.0 & 63.3 & 60.8 & 58.9 & 62.9 & 59.1 & 50.7 & 46.6 & 45.8 & 50.0 & 44.4 & 44.4 & 44.2 & 46.8 & 38.5 \\
\hline
\multirow{3}{*}{\rotatebox[origin=c]{90}{IN-Sk.}}
& Source & 80.9 & 81.6 & 76.4 & 78.7 & 78.1 & 79.6 & 77.7 & 72.8 & 69.0 & 68.5 & 74.0 & 69.4 & 69.3 & 70.6 & 72.2 & 65.1 \\
& BN--1  & \textcolor{red}{81.4} & \textcolor{red}{86.9} & \textcolor{red}{82.0} & \textcolor{red}{80.4} & \textcolor{red}{79.2} & \textcolor{red}{81.8} & \textcolor{red}{79.5} &   -  &   -  &   -  &   -  &   -  &   -  &   -  &   -  & 67.0 \\
& ROID   & 74.2 & \textcolor{red}{83.8} & \textcolor{red}{77.6} & 75.6 & 73.1 & 76.5 & 74.0 & 63.5 & 59.6 & 58.6 & 64.0 & 58.9 & 58.7 & 58.6 & 59.9 & 55.2 \\
\hline
\multirow{3}{*}{\rotatebox[origin=c]{90}{IN-D109}}
& Source & 62.5 & 63.5 & 59.5 & 60.0 & 59.4 & 60.1 & 58.6 & 54.3 & 51.8 & 51.4 & 55.2 & 51.8 & 51.5 & 53.6 & 55.9 & 49.4 \\
& BN--1  & 60.6 & \textcolor{red}{66.0} & \textcolor{red}{61.8} & \textcolor{red}{60.8} & \textcolor{red}{59.6} & \textcolor{red}{62.1} & \textcolor{red}{59.9} &   -  &   -  &   -  &   -  &   -  &   -  &   -  &   -  & 48.8 \\
& ROID   & 55.1 & 62.6 & 58.6 & 55.6 & 54.4 & 57.8 & 55.1 & 48.1 & 45.6 & 45.0 & 48.6 & 45.0 & 44.3 & 45.0 & 47.1 & 41.9 \\
\hline
\end{tabular}
}
\end{table*}

To demonstrate that our proposed method ROID is largely model-agnostic, we evaluate our method in the continual TTA setting on 31 different architectures. In \Cref{tab:architectures-regular}, we report our results on regular architectures. In \Cref{tab:architectures-mobile}, mobile architectures and transformers are considered on the left and right, respectively. All results worse than the source performance are highlighted in red. While test-time normalization (BN--1) can decrease the error for corruptions (IN-C) on all considered architectures, this is not the case for natural shifts (IN-R, IN-Sketch, IN-D109). Especially for mobile architectures, Inception-v3, and RegNets, the error rate even increases. Since ROID applies test-time normalization, it works particularly well when a good estimation of the batch statistics is possible during test-time. ROID always outperforms BN--1, but due to the bad estimation of the batch statistics of MobileNet-v3 on ImageNet-Sketch, improvement upon the source performance is not possible. Nevertheless, in general, ROID can significantly outperform the source model, demonstrating its applicability to a wide range of different architectures. Among all networks, MaxViT-tiny, a hybrid (CNN + ViT) model, performs best on all ImageNet benchmarks. Regarding the considered CNN architectures, ResNeXt-101-32x8d and RegNetY-32gf show the best overall results.

\subsection{Catastrophic Forgetting}
\label{sec:appendix_forgetting}
\begin{figure}[h]
    \centering
    \includegraphics[scale=0.5]{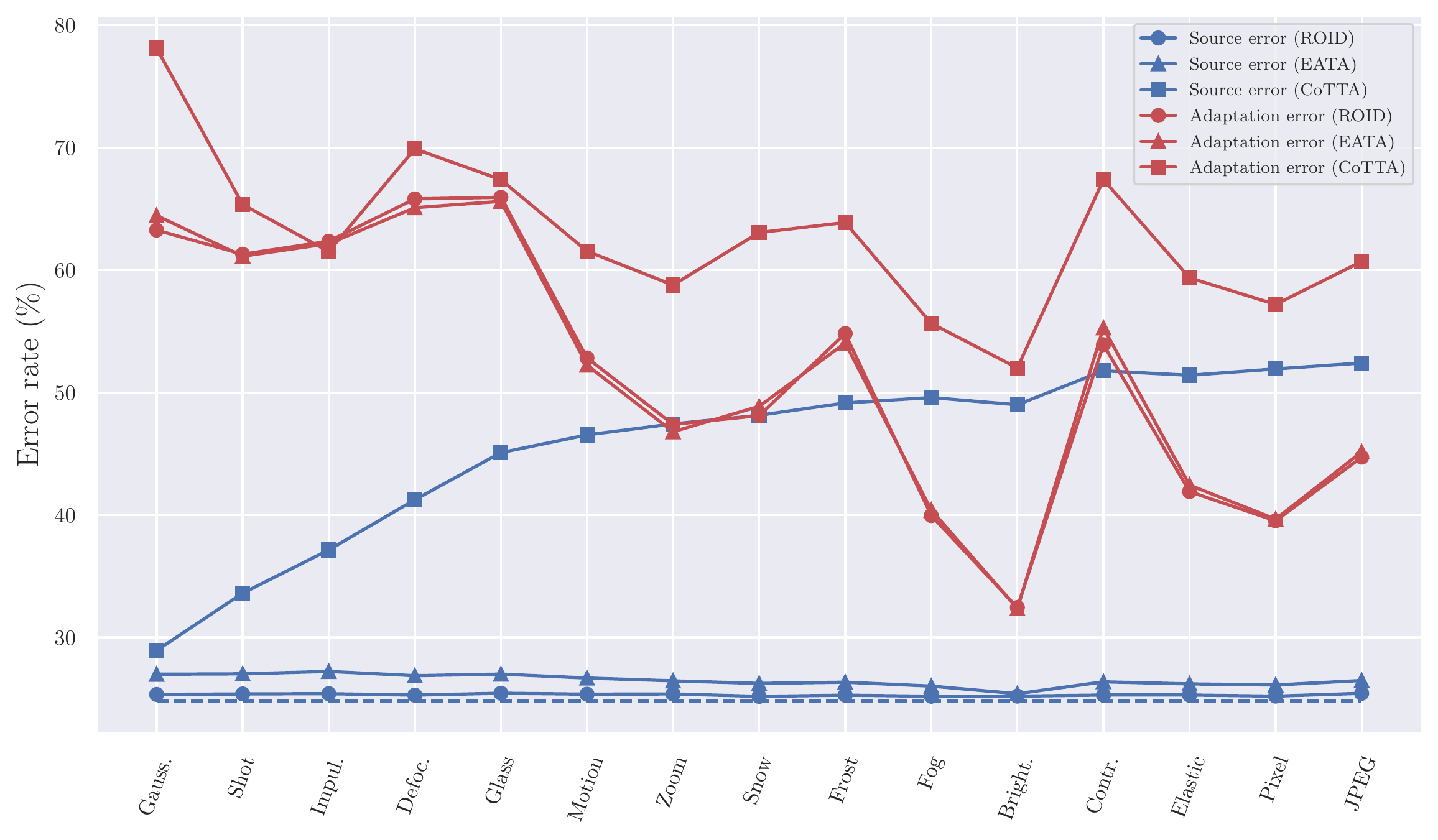}
    \vskip -0.1in
    \caption{Source and adaptation error of ROID, EATA, and CoTTA for ImageNet-C (50,000 samples per domain) in the continual TTA setting with an alternating domain sequence. The dashed line indicates the lower bound (source error of the source model).}
    \label{fig:forgetting}
    \vskip -0.2in
\end{figure}

In \Cref{fig:forgetting}, we investigate the occurrence of catastrophic forgetting \cite{mccloskey1989catastrophic} for CoTTA \cite{wang2022continual}, EATA \cite{niu2022efficient}, and ROID on the long continual ImageNet-C sequence (50,000 samples per corruption). Following \cite{niu2022efficient}, we adapt the model on an alternating sequence of corrupted data and source data, i.e., [\textit{Gaussian}, \textit{Source}, \textit{Shot}, \textit{Source}, ...], using the complete ImageNet validation set (50,000 samples) as \textit{Source}. Note that this procedure is different compared to how catastrophic forgetting is measured within the field of continual learning. However, in TTA, where the model is continually adapted to an unknown domain, this is the more realistic setting. Clearly, CoTTA suffers from major catastrophic forgetting, as the source error steadily increases after each corruption. By using elastic weight consolidation, EATA can largely prevent forgetting. However, to perform elastic weight consolidation, EATA requires data from the initial source domain, which may be unavailable in practice. Our proposed method ROID, which utilizes weight ensembling, is even more effective than EATA and only requires the initial parameters of the normalization layers. ROID is capable of nearly recovering the performance of the initial source model on the source domain.

\subsection{Momentum for Weight Ensembling}
In \Cref{tab:appendix_momentum}, we analyze the sensitivity with respect to the momentum $\alpha$ used for our weight ensembling. ResNet-50, Swin-b, and ViT-b-16 are evaluated on the continual ImageNet-C benchmark. Choosing a relatively low momentum $\alpha=0.9$, corresponding to only "keeping" 90\% of the current model and adding 10\% of the weights of the initial source model, limits adaptation. In the interval $\alpha\in[0.99, 0.9975]$, a decent compromise between allowing adaptation and remaining good generalization from the source model is possible. For large momentum values $\alpha\geq0.999$ the advantages of weight ensembling vanish, resulting in an increase of adaptation error for all architectures.
\begin{table}[h]
\renewcommand{\arraystretch}{1.2}
\centering
\caption{Online classification error rate~(\%) for ImageNet-C at the highest severity level 5 in the \textit{continual} TTA setting. Different momentum values used for weight ensembling are considered for our approach. Note that we omitted prior correction and $\mathcal{L}_{\mathrm{SCE}}$ for a clearer analysis.} \label{tab:appendix_momentum}
\vskip -0.1in
\scalebox{0.9}{
\tabcolsep4pt
\begin{tabular}{l|cccccccc}\hline
\diagbox{Model}{$\alpha$} & 0.99975 & 0.9995 & 0.999 & 0.9975 & 0.995 & 0.99 & 0.95 & 0.9  \\
\hline
ResNet-50   & 60.1 & 58.9 & 58.4 & 57.0 & 56.3 & \textbf{56.1} & 60.0 & 63.0 \\
Swin-b      & 53.4 & 52.4 & 51.9 & 50.8 & 50.1 & \textbf{49.7} & 53.4 & 57.0 \\
ViT-b-16    & 47.9 & 47.8 & 47.4 & 46.7 & \textbf{46.7} & 47.0 & 51.8 & 55.3 \\
\hline
\end{tabular}
}
\end{table}

\subsection{Computational Efficiency}
Since efficiency is also of great importance for a method performing its adaptation during test-time, we study in \Cref{tab:appendix_efficiency_compute} the efficiency of each method with respect to the number of required forward and backward propagations, as well as the number of trainable parameters. We conduct the analysis on ImageNet-R using a ResNet-50. Clearly, the most inefficient methods are CoTTA, RMT, and AdaContrast which do not only require three and four times as many forward passes, but also calculate the gradients with respect to all parameters. While RoTTA also performs three forward passes per test sample, significantly less parameters are trained and the number of backward propagations is not increased. The most efficient method during adaptation is EATA. Compared to the second most efficient method, TENT, fewer backward passes are required as some samples are filtered out. Due to performing consistency regularization, ROID is slightly less efficient than TENT and EATA, but comparable to SAR. Note that the additional 2000 forward and backward passes required to calculate the Fisher information matrix in EATA are not included in \Cref{tab:appendix_efficiency_compute}.
\begin{table}[h]
\renewcommand{\arraystretch}{1.2}
\centering
\caption{Efficiency analysis for adapting a ResNet-50 on ImageNet-R.} \label{tab:appendix_efficiency_compute}
\vskip -0.1in
\scalebox{0.9}{
\tabcolsep4pt
\begin{tabular}{l|cccc}\hline
Method  & Error $(\%)$ & \#Forwards & \#Backwards & Train. Params $(\%)$ \\
\hline
Source      & 63.8 & 30,000 & - & - \\
BN--1       & 60.3 & 30,000 & - & - \\
LAME        & 99.4 & 30,000 & - & - \\
\hline
TENT-cont.  & 57.4 & 30,000 & 30,000 & 0.21 \\
EATA        & 54.2 & 30,000 &  5,440 & 0.21 \\
SAR         & 57.2 & 46,279 & 30,111 & 0.12 \\
CoTTA       & 57.4 & 90,000 & 30,000 & 100 \\
RoTTA       & 60.8 & 90,000 & 30,000 & 0.21 \\
AdaContrast & 59.1 & 120,000 & 60,000 & 100 \\ 
RMT         & 55.9 & 90,000 & 60,000 & 100 \\
ROID (ours) & 51.3 & 48,610 & 37,220 & 0.21 \\
\hline
\end{tabular}
}
\end{table}

\subsection{Memory Efficiency}
Another huge advantage of architectures based on group or layer normalization is their potential to recover the batch TTA setting from a single sample scenario by leveraging gradient accumulation. This approach has the additional benefit that it significantly reduces the amount of required memory, which can be a scarce when TTA is performed on an edge device. In \Cref{tab:appendix_efficiency_memory}, the allocated memory for the batch and single sample setting is compared. Using gradient accumulation with TENT and ViT-b-16 reduces the maximum GPU memory consumption by 14.5 times while providing the same results. In case of ROID, the reduction factor is 15.8. If Swin-b is used as a model, the memory reduction factors are even larger.
\begin{table}[h]
\renewcommand{\arraystretch}{1.2}
\centering
\caption{Memory efficiency analysis for TENT-cont. and ROID when adapting either Swin-b or ViT-b-16 on ImageNet-R.} \label{tab:appendix_efficiency_memory}
\vskip -0.1in
\scalebox{0.9}{
\tabcolsep4pt
\begin{tabular}{l|cccc}\hline
Method  & Architecture & Batch Size & Error $(\%)$ & Max. GPU mem. allocated \\
\hline
TENT-cont. & Swin-b & 64 & 54.2 & 9.20 GB \\
TENT-cont. & Swin-b & 1 & 54.3 & 0.50 GB \\
TENT-cont. & ViT-b-16 & 64 & 53.3 & 6.36 GB \\
TENT-cont. & ViT-b-16 & 1 & 53.3 & 0.44 GB \\
\hline
ROID (ours) & Swin-b & 64 & 45.8 & 15.92 GB \\
ROID (ours) & Swin-b & 1 & 45.8 & 0.71 GB \\
ROID (ours) & ViT-b-16 & 64 & 44.2 & 10.90 GB \\
ROID (ours) & ViT-b-16 & 1 & 44.1 & 0.69 GB \\
\hline
\end{tabular}
}
\end{table}

\subsection{Component Analysis}
\label{sec:appendix_component_analysis}
In the following we elaborate and extend the component analysis from the main paper. Detailed results for the continual and mixed-domains setting are presented in \Cref{tab:ablation-components-continual} and for the correlated and mixed-domains correlated setting in \Cref{tab:ablation-components-correlated}. To adapt a model to the entire spectrum of Universal TTA, the most important aspect is to have a stable method. This factor isn't solely crucial for a specific scenario in TTA but resonates across all settings. As our analysis in \cref{sec:self_training_analysis} and \cref{sec:appendix_error_accumulation} suggests, even in the easiest setting (continual) it is essential to prevent the model from developing a bias or worse, collapsing to a trivial solution during test-time. A non-stationary setting, such as mixed-domains, can further enhance a model bias and degrade performance. To circumvent this, diversity weighting is essential. This is also supported by our component analysis which demonstrates that the driving factor in the continual and mixed-domains setting is diversity and certainty weighting.

To effectively address the challenge of dealing with multiple domain shifts over time, we employ weight ensembling (WE). WE retains generalization and still enables a good adaptation, as demonstrated in \cref{sec:self_training_analysis}. It should be underscored that this is only necessary when a model adapts to a narrow distribution, potentially leading to overfitting on the current domain. In the context of mixed-domains, where samples from different domains are encountered within a single batch, adapting to such a broad distribution is also possible without WE. This is demonstrated by our component analysis, where WE improves the performance where multiple domain shifts are encountered, but actually slightly degrades the performance in the mixed-domains setting (broad distribution). Note that also for ImageNet-R and ImageNet-Sketch the best performance in the continual setting is achieved for configuration B, since here we only adapt to a single domain and do not encounter any additional domain shifts where generalization would be of importance. Nevertheless, the concept of WE carries the added benefit of enhancing overall stability. It serves as a corrective measure, capable of rectifying suboptimal adaptations over time, by continually incorporating a small percentage of the source weights. This becomes visible for the difficult adaptation in correlated settings, where highly imbalanced data can hinder a stable adaptation process. Here, WE ensembling ensures a stable adaptation process.

Shifting our focus to the correlated setting, the role of prior correction is substantial. Weighting the network's outputs with a smoothed estimate of the label prior benefits in settings with highly imbalanced data. Uncertain data points can be corrected by taking prior label information into account, while not degrading performance when a uniform label distribution is present.

Taking a look at employing consistency through data augmentation, the component analysis shows that it is beneficial across all settings and datasets. Compared to the other components, encouraging the invariance to small changes in the input space, has a moderate benefit.

\newpage
\section{Detailed Results}
\label{sec:appendix_detailed_results}
\begin{table*}[!h]
\renewcommand{\arraystretch}{1.125}
\centering
\caption{Online classification error rate~(\%) for different settings using the ImageNet-D109 dataset. We report the performance of each method averaged over 5 runs. We do not report the results for ResNet-50 in the correlated setting, since BN--1 already achieves an error of 92.8\%.} \label{tab:detailed-imagenet-d109}
\vskip -0.1in
\scalebox{0.79}{
\tabcolsep4pt
\begin{tabular}{l|l|ccccc|c||ccccc|c||ccccc|c}\hline
& Setting & \multicolumn{5}{c|}{continual} & & \multicolumn{5}{c|}{correlated} & & \multicolumn{5}{c|}{mixed domains} & \\ \hline
& Time & \multicolumn{5}{l|}{$t\xrightarrow{\hspace*{3.7cm}}$} &  & \multicolumn{5}{l|}{$t\xrightarrow{\hspace*{3.7cm}}$} &  &  &  &  &  &  \\ \hline
& Method & \rotatebox[origin=c]{70}{clipart} & \rotatebox[origin=c]{70}{infograph} & \rotatebox[origin=c]{70}{painting} & \rotatebox[origin=c]{70}{real} & \rotatebox[origin=c]{70}{sketch} & Mean &\rotatebox[origin=c]{70}{clipart} & \rotatebox[origin=c]{70}{infograph} & \rotatebox[origin=c]{70}{painting} & \rotatebox[origin=c]{70}{real} & \rotatebox[origin=c]{70}{sketch} & Mean & \rotatebox[origin=c]{70}{clipart} & \rotatebox[origin=c]{70}{infograph} & \rotatebox[origin=c]{70}{painting} & \rotatebox[origin=c]{70}{real} & \rotatebox[origin=c]{70}{sketch} & Mean\\
\hline
\multirow{11}{*}{\rotatebox[origin=c]{90}{ ResNet-50 }} 
& Source & 64.2 & 81.0 & 51.5 & 24.2 & 73.2 & 58.8 & 64.2 & 81.0 & 51.5 & 24.2 & 73.2 & 58.8 & 64.2 & 81.0 & 51.5 & 24.2 & 73.2 & 58.8 \\
& BN--1        & 55.7 & 80.1 & 50.1 & \textcolor{red}{25.1} & 64.8 & 55.1$\pm$0.05 & \textcolor{red}{92.4} & \textcolor{red}{93.2} & \textcolor{red}{91.9} & \textcolor{red}{92.7} & \textcolor{red}{94.0} & \textcolor{red}{92.8}$\pm$0.05 & 58.3 & 78.6 & \textcolor{red}{51.7} & \textcolor{red}{26.0} & 66.6 & 56.2$\pm$0.03 \\
& TENT-c.   & 53.5 & 78.1 & 47.9 & \textcolor{red}{24.8} & 60.3 & 52.9$\pm$0.05 & - & - & - & - & - & - & 57.8 & \textcolor{red}{81.7} & 50.1 & \textcolor{red}{25.3} & 65.7 & 56.1$\pm$0.15 \\
& EATA         & 51.8 & 76.7 & 47.6 & 24.0 & 57.8 & 51.6$\pm$0.21 & - & - & - & - & - & - & 54.2 & 78.4 & 49.3 & \textcolor{red}{24.3} & 60.5 & 53.3$\pm$0.3 \\
& SAR          & 53.9 & 77.6 & 47.3 & \textcolor{red}{24.4} & 58.1 & 52.2$\pm$0.07 & - & - & - & - & - & - & 55.2 & 77.5 & 49.1 & \textcolor{red}{24.8} & 61.9 & 53.7$\pm$0.05 \\
\hline
\end{tabular}
}
\end{table*}

\begin{table}[!h]
\renewcommand{\arraystretch}{1.1}
\centering
\caption{Online classification error rate~(\%) for ImageNet-D109 for the \textit{mixed domains correlated} TTA setting with Dirichlet concentration parameter $\delta=0.1$. We report the performance of each method averaged over 5 runs.} \label{tab:detailed-imagenet-d109-mixed-correlated}
\vskip -0.1in
\scalebox{0.8}{
\tabcolsep5pt
\begin{tabular}{l|l|ccccc|c}\hline
& Method & \rotatebox[origin=c]{70}{clipart} & \rotatebox[origin=c]{70}{infograph} & \rotatebox[origin=c]{70}{painting} & \rotatebox[origin=c]{70}{real} & \rotatebox[origin=c]{70}{sketch} & Mean \\
\hline
\multirow{4}{*}{\rotatebox[origin=c]{90}{Swin-b}} & Source & 53.6 & 73.6 & 44.0 & 20.3 & 65.3 & 51.4 \\
& SAR & \textcolor{red}{56.4} & \textcolor{red}{78.0} & \textcolor{red}{46.5} & \textcolor{red}{21.2} & \textcolor{red}{67.4} & \textcolor{red}{53.9}$\pm$0.52 \\
& LAME & 26.6 & \textbf{25.9} & 30.8 & \textcolor{red}{28.2} & \textbf{28.3} & \textbf{28.0}$\pm$0.39 \\
& ROID & \textbf{25.5} & 47.0 & \textbf{24.0} & \textbf{12.4} & 32.6 & 28.3$\pm$0.19 \\
\hline
\multirow{4}{*}{\rotatebox[origin=c]{90}{ViT-b-16}} & Source & 57.5 & 75.9 & 45.1 & 22.0 & 67.5 & 53.6 \\
& SAR & \textcolor{red}{66.3} & \textcolor{red}{82.4} & \textcolor{red}{53.6} & \textcolor{red}{25.5} & \textcolor{red}{76.4} & \textcolor{red}{60.8}$\pm$0.48 \\
& LAME & \textbf{27.6} & \textbf{27.3} & 32.2 & \textcolor{red}{29.6} & \textbf{29.4} & \textbf{29.2}$\pm$0.55 \\
& ROID & 28.0 & 47.8 & \textbf{24.7} & \textbf{12.8} & 33.8 & 29.4$\pm$0.13 \\
\hline
\end{tabular}}
\end{table}
\begin{table}[H]
\renewcommand{\arraystretch}{1.1}
\centering
\caption{Online classification error rate~(\%) for the corruption benchmarks at the highest severity level 5 for the \textit{continual} TTA setting. For CIFAR10-C the results are evaluated on WideResNet-28, for CIFAR100-C on ResNeXt-29, and for Imagenet-C, ResNet-50, Swin-b and ViT-b-16 are used. We report the performance of each method averaged over 5 runs.} \label{tab:detailed-continual}
\vskip -0.1in
\scalebox{0.8}{
\tabcolsep5pt
\begin{tabular}{l|l|ccccccccccccccc|c}\hline
& Time & \multicolumn{15}{l|}{$t\xrightarrow{\hspace*{13.7cm}}$}& \\ \hline
& Method & \rotatebox[origin=c]{70}{Gaussian} & \rotatebox[origin=c]{70}{shot} & \rotatebox[origin=c]{70}{impulse} & \rotatebox[origin=c]{70}{defocus} & \rotatebox[origin=c]{70}{glass} & \rotatebox[origin=c]{70}{motion} & \rotatebox[origin=c]{70}{zoom} & \rotatebox[origin=c]{70}{snow} & \rotatebox[origin=c]{70}{frost} & \rotatebox[origin=c]{70}{fog}  & \rotatebox[origin=c]{70}{ bright. } & \rotatebox[origin=c]{70}{contrast} & \rotatebox[origin=c]{70}{elastic} & \rotatebox[origin=c]{70}{pixelate} & \rotatebox[origin=c]{70}{jpeg} & Mean \\
\hline
\multirow{11}{*}{\rotatebox[origin=c]{90}{CIFAR10-C}} & Source & 72.3 & 65.7 & 72.9 & 47.0 & 54.3 & 34.8 & 42.0 & 25.1 & 41.3 & 26.0 & 9.3 & 46.7 & 26.6 & 58.4 & 30.3 & 43.5 \\
& BN--1 & 28.3 & 26.2 & 36.2 & 12.7 & 35.1 & 13.9 & 12.2 & 17.5 & 17.7 & 15.0 & 8.3 & 13.0 & 23.6 & 19.7 & 27.4 & 20.4$\pm$0.07 \\
& TENT-cont. & 25.0 & 20.3 & 29.0 & 13.8 & 31.7 & 16.2 & 14.1 & 18.6 & 17.6 & 17.4 & \textcolor{red}{10.8} & 15.6 & 24.3 & 19.7 & 25.1 & 20.0$\pm$1.19 \\
& EATA & 24.6 & 19.1 & 27.7 & 12.8 & 29.4 & 14.5 & 12.1 & 16.3 & 15.8 & 15.2 & 9.3 & 13.0 & 21.6 & 16.1 & 20.8 & 17.9$\pm$0.15 \\
& SAR & 28.3 & 26.2 & 35.6 & 12.7 & 34.7 & 13.9 & 12.2 & 17.5 & 17.7 & 15.0 & 8.3 & 13.0 & 23.6 & 19.7 & 27.4 & 20.4$\pm$0.06 \\
& CoTTA & 24.2 & 21.9 & 26.5 & 12.0 & 27.9 & 12.7 & 10.7 & 15.2 & 14.6 & 12.8 & 7.9 & 11.2 & 18.5 & \textbf{14.0} & 18.1 & 16.5$\pm$0.16 \\
\hline
\end{tabular}
}
\end{table}
\begin{table*}[!h]
\renewcommand{\arraystretch}{1.1}
\centering
\caption{Online classification error rate~(\%) for the corruption benchmarks at the highest severity level 5 for the \textit{mixed domains} TTA setting. For CIFAR10-C the results are evaluated on WideResNet-28, for CIFAR100-C on ResNeXt-29, and for Imagenet-C, ResNet-50, Swin-b and ViT-b-16 are used. We report the performance of each method averaged over 5 runs.} \label{tab:detailed-mixed}
\vskip -0.1in
\scalebox{0.8}{
\tabcolsep5pt
\begin{tabular}{l|l|ccccccccccccccc|c}\hline
& Method & \rotatebox[origin=c]{70}{Gaussian} & \rotatebox[origin=c]{70}{shot} & \rotatebox[origin=c]{70}{impulse} & \rotatebox[origin=c]{70}{defocus} & \rotatebox[origin=c]{70}{glass} & \rotatebox[origin=c]{70}{motion} & \rotatebox[origin=c]{70}{zoom} & \rotatebox[origin=c]{70}{snow} & \rotatebox[origin=c]{70}{frost} & \rotatebox[origin=c]{70}{fog}  & \rotatebox[origin=c]{70}{ bright. } & \rotatebox[origin=c]{70}{contrast} & \rotatebox[origin=c]{70}{elastic} & \rotatebox[origin=c]{70}{pixelate} & \rotatebox[origin=c]{70}{jpeg} & Mean \\
\hline
\multirow{11}{*}{\rotatebox[origin=c]{90}{CIFAR10-C}} & Source & 72.3 & 65.7 & 72.9 & 46.9 & 54.3 & 34.8 & 42.0 & 25.1 & 41.3 & 26.0 & 9.3 & 46.7 & \textbf{26.6} & 58.4 & 30.3 & 43.5 \\
& BN--1 & 45.5 & 42.8 & 59.7 & 34.2 & 44.3 & 29.8 & 32.0 & 19.8 & 21.1 & 21.5 & 9.3 & 27.9 & \textcolor{red}{33.1} & 55.5 & \textcolor{red}{30.8} & 33.8$\pm$0.04 \\
& TENT-cont. & \textcolor{red}{73.5} & \textcolor{red}{70.1} & \textcolor{red}{81.4} & 31.6 & \textcolor{red}{60.3} & 29.6 & 28.5 & \textcolor{red}{30.8} & 35.3 & 25.7 & \textcolor{red}{13.6} & 44.2 & \textcolor{red}{32.6} & \textcolor{red}{70.2} & \textcolor{red}{34.9} & \textcolor{red}{44.1}$\pm$3.82 \\
& EATA & \textbf{36.4} & \textbf{33.5} & 51.5 & 24.1 & 38.9 & 23.4 & 21.5 & 19.8 & 19.8 & 21.5 & \textcolor{red}{11.4} & 32.0 & \textcolor{red}{27.1} & 42.2 & 25.3 & 28.6$\pm$0.7 \\
& SAR & 45.5 & 42.7 & 59.6 & 34.1 & 44.3 & 29.7 & 31.9 & 19.8 & 21.1 & 21.5 & 9.3 & 27.8 & \textcolor{red}{33.0} & 55.4 & \textcolor{red}{30.8} & 33.8$\pm$0.04 \\
& CoTTA & 38.7 & 36.0 & 56.1 & 36.0 & \textbf{36.8} & 32.3 & 31.0 & 19.9 & \textbf{17.6} & \textcolor{red}{27.2} & \textcolor{red}{11.7} & \textcolor{red}{52.6} & \textcolor{red}{30.5} & \textbf{35.8} & 25.7 & 32.5$\pm$1.35 \\
\hline
\end{tabular}
}
\end{table*}
\begin{table*}[t]
\renewcommand{\arraystretch}{1.2}
\centering
\caption{Online classification error rate~(\%) in the \textit{correlated} TTA setting where samples are sorted by class. The corruption datasets are evaluated at the highest severity level 5. We report the performance of each method averaged over 5 runs.} \label{tab:correlated}
\vskip -0.1in
\scalebox{0.9}{
\tabcolsep4pt
\begin{tabular}{l|l|ccccccccc|c}\hline
Dataset & Architecture & Source & TENT & EATA & SAR & CoTTA & RoTTA & AdaCont. & RMT & LAME & ROID (ours) \\
\hline
CIFAR10-C & RN-26 GN & 32.7 & \textcolor{red}{87.6} & \textcolor{red}{40.8} & \textcolor{red}{37.1} & \textcolor{red}{44.5} & \textcolor{red}{33.7} & 30.5 & \textcolor{red}{57.5} & \textbf{11.3} & 15.9$\pm$0.27 \\
\hline
\multirow{2}{*}{IN-C} & Swin-b & 64.0 & \textcolor{red}{86.7} & \textcolor{red}{74.2} & 59.3 & \textcolor{red}{99.5} & \textcolor{red}{75.5} & \textcolor{red}{77.6} & \textcolor{red}{99.6} & 47.0 & \textbf{18.5}$\pm$0.10 \\
& ViT-b-16 & 60.2 & \textcolor{red}{80.6} & \textcolor{red}{76.2} & 53.9 & \textcolor{red}{98.8} & \textcolor{red}{65.1} & \textcolor{red}{87.4} & \textcolor{red}{99.6} & 44.1 & \textbf{16.8}$\pm$0.72 \\
\hline
\multirow{2}{*}{IN-R} & Swin-b & 54.2 & 53.6 & 53.9 & 53.1 & \textcolor{red}{58.9} & 54.1 & \textcolor{red}{56.9} & 48.1 & \textbf{13.6} & 25.2$\pm$0.37 \\
& ViT-b-16 & 56.0 & 53.4 & 53.6 & 49.9 & \textcolor{red}{81.0} & 55.8 & \textcolor{red}{62.1} & \textcolor{red}{85.8} & \textbf{13.0} & 25.8$\pm$0.13 \\
\hline
\multirow{2}{*}{IN-Sketch} & Swin-b & 68.4 & 67.4 & 66.3 & \textcolor{red}{72.3} & \textcolor{red}{95.3} & 68.1 & 66.9 & \textcolor{red}{91.8} & 58.2 & \textbf{43.9}$\pm$0.19 \\
& ViT-b-16 & 70.6 & 66.7 & 63.7 & \textcolor{red}{74.6} & \textcolor{red}{95.5} & 70.1 & \textcolor{red}{72.3} & \textcolor{red}{97.9} & 61.0 & \textbf{44.0}$\pm$0.14 \\
\hline
\multirow{2}{*}{IN-D109} & Swin-b & 51.4 & \textcolor{red}{68.5} & \textcolor{red}{53.9} & \textcolor{red}{55.5} & \textcolor{red}{58.5} & 50.5 & \textcolor{red}{52.1} & \textcolor{red}{51.9} & \textbf{30.4} & 30.6$\pm$0.16 \\
& ViT-b-16 & 53.6 & \textcolor{red}{84.3} & \textcolor{red}{57.4} & \textcolor{red}{58.7} & \textcolor{red}{93.1} & \textcolor{red}{53.8} & \textcolor{red}{56.7} & \textcolor{red}{90.6} & 35.4 & \textbf{31.7}$\pm$0.08 \\
\hline
\end{tabular}
}
\vskip -0.1in
\end{table*}

\begin{table}[H]
\renewcommand{\arraystretch}{1.2}
\centering
\caption{Online classification error rate~(\%) for the corruption benchmarks at the highest severity level 5 for the \textit{correlated} TTA setting. For CIFAR10-C the results are evaluated on ResNet-26 with group norm (RN-26 GN). For Imagenet-C, Swin-b, and ViT-b-16 are used. We report the performance of each method averaged over 5 runs.} \label{tab:detailed-correlated}
\vskip -0.1in
\scalebox{0.9}{
\tabcolsep4pt
\begin{tabular}{l|l|ccccccccccccccc|c}\hline
& Time & \multicolumn{15}{l|}{$t\xrightarrow{\hspace*{12.8cm}}$}& \\ \hline
& Method & \rotatebox[origin=c]{70}{Gaussian} & \rotatebox[origin=c]{70}{shot} & \rotatebox[origin=c]{70}{impulse} & \rotatebox[origin=c]{70}{defocus} & \rotatebox[origin=c]{70}{glass} & \rotatebox[origin=c]{70}{motion} & \rotatebox[origin=c]{70}{zoom} & \rotatebox[origin=c]{70}{snow} & \rotatebox[origin=c]{70}{frost} & \rotatebox[origin=c]{70}{fog}  & \rotatebox[origin=c]{70}{ bright. } & \rotatebox[origin=c]{70}{contrast} & \rotatebox[origin=c]{70}{elastic} & \rotatebox[origin=c]{70}{pixelate} & \rotatebox[origin=c]{70}{jpeg} & Mean \\
\hline
\multirow{10}{*}{\rotatebox[origin=c]{90}{CIFAR10-C (RN-26 GN)}}
& Source & 48.4 & 44.8 & 50.3 & 24.1 & 47.8 & 24.5 & 24.1 & 24.1 & 33.1 & 28.0 & 14.1 & 29.7 & 25.6 & 43.7 & 28.3 & 32.7 \\
& TENT-cont. & \textcolor{red}{62.3} & \textcolor{red}{82.6} & \textcolor{red}{89.9} & \textcolor{red}{89.4} & \textcolor{red}{90.0} & \textcolor{red}{90.0} & \textcolor{red}{90.0} & \textcolor{red}{90.0} & \textcolor{red}{90.0} & \textcolor{red}{90.0} & \textcolor{red}{90.0} & \textcolor{red}{90.0} & \textcolor{red}{90.0} & \textcolor{red}{90.0} & \textcolor{red}{90.0} & \textcolor{red}{87.6}$\pm$0.98 \\
& EATA & 39.3 & 30.9 & 42.6 & \textcolor{red}{30.3} & 45.4 & \textcolor{red}{24.9} & \textcolor{red}{33.6} & \textcolor{red}{36.7} & \textcolor{red}{36.1} & \textcolor{red}{41.4} & \textcolor{red}{36.0} & \textcolor{red}{47.5} & \textcolor{red}{52.6} & \textcolor{red}{62.9} & \textcolor{red}{51.0} & \textcolor{red}{40.8}$\pm$1.62 \\
& SAR & \textcolor{red}{48.6} & \textcolor{red}{55.6} & \textcolor{red}{63.8} & 23.4 & \textcolor{red}{55.8} & \textcolor{red}{29.1} & \textcolor{red}{24.9} & \textcolor{red}{30.6} & 33.1 & 26.1 & \textcolor{red}{14.5} & 27.8 & \textcolor{red}{31.6} & \textcolor{red}{58.5} & \textcolor{red}{33.0} & \textcolor{red}{37.1}$\pm$1.12 \\
\hline
\end{tabular}
}
\end{table}

\begin{table*}[!h]
\renewcommand{\arraystretch}{1.2}
\centering
\caption{Online classification error rate~(\%) for ImageNet-C at the highest severity level 5 for the \textit{mixed domains correlated} TTA setting with the Dirichlet concentration parameter $\delta=0.01$. We report the performance of each method averaged over 5 runs.} \label{tab:detailed-mixed-correlated}
\vskip -0.1in
\scalebox{0.9}{
\tabcolsep4pt
\begin{tabular}{l|l|ccccccccccccccc|c}\hline
& Method & \rotatebox[origin=c]{70}{Gaussian} & \rotatebox[origin=c]{70}{shot} & \rotatebox[origin=c]{70}{impulse} & \rotatebox[origin=c]{70}{defocus} & \rotatebox[origin=c]{70}{glass} & \rotatebox[origin=c]{70}{motion} & \rotatebox[origin=c]{70}{zoom} & \rotatebox[origin=c]{70}{snow} & \rotatebox[origin=c]{70}{frost} & \rotatebox[origin=c]{70}{fog}  & \rotatebox[origin=c]{70}{ bright. } & \rotatebox[origin=c]{70}{contrast} & \rotatebox[origin=c]{70}{elastic} & \rotatebox[origin=c]{70}{pixelate} & \rotatebox[origin=c]{70}{jpeg} & Mean \\
\hline
\multirow{4}{*}{\rotatebox[origin=c]{90}{Swin-b}} & Source & 70.4 & 69.9 & 75.5 & 72.8 & 81.9 & 64.4 & 68.6 & 57.9 & 50.5 & 40.7 & 29.2 & 59.8 & 72.6 & 87.0 & 58.8 & 64.0 \\
& SAR & \textcolor{red}{70.8} & 68.8 & 71.2 & \textcolor{red}{73.4} & 76.2 & 61.4 & 66.0 & \textcolor{red}{65.3} & \textcolor{red}{61.1} & \textcolor{red}{56.7} & \textcolor{red}{32.8} & \textcolor{red}{62.0} & \textcolor{red}{73.2} & 79.6 & 55.7 & \textcolor{red}{64.9}$\pm$0.81 \\
& LAME & 37.4 & 37.4 & 37.4 & 37.6 & 37.8 & 37.4 & 37.7 & 37.2 & 37.0 & 37.3 & \textcolor{red}{36.5} & 37.5 & 37.6 & 37.8 & 37.2 & 37.4$\pm$0.12 \\
& ROID (ours) & \textbf{30.9} & \textbf{30.1} & \textbf{31.0} & \textbf{34.8} & \textbf{37.2} & \textbf{26.4} & \textbf{31.1} & \textbf{26.8} & \textbf{22.8} & \textbf{23.2} & \textbf{12.3} & \textbf{27.0} & \textbf{33.6} & \textbf{36.4} & \textbf{25.2} & \textbf{28.6}$\pm$0.16 \\
\hline
\multirow{4}{*}{\rotatebox[origin=c]{90}{ViT-b-16}} & Source & 66.0 & 66.8 & 64.9 & 68.5 & 74.7 & 64.0 & 66.9 & 57.3 & 45.0 & 49.4 & 28.7 & 81.8 & 57.8 & 60.8 & 49.9 & 60.2 \\
& SAR & 64.5 & 62.9 & 63.2 & 59.0 & 64.1 & 52.8 & 60.6 & 54.8 & \textcolor{red}{49.6} & 47.4 & \textcolor{red}{30.6} & 59.1 & 53.5 & 49.0 & 42.9 & 54.3$\pm$0.59 \\
& LAME & 36.2 & 36.1 & 36.1 & 36.3 & 36.3 & 36.1 & 36.4 & 36.1 & 35.8 & 36.1 & \textcolor{red}{35.4} & 36.6 & 35.9 & 36.0 & 35.7 & 36.1$\pm$0.15 \\
\hline
\end{tabular}
}
\end{table*}

\begin{table}[h]
\renewcommand{\arraystretch}{1.2}
\centering
\caption{Average online classification error rate~(\%) over 5 runs for different configurations for a) the \textit{continual} TTA setting and b) the \textit{mixed domains} TTA setting. For the ImageNet variants, a ResNet-50 is used. For CIFAR10-C and CIFAR100-C, the results are evaluated utilizing a WideResNet-28 and a ResNeXt-29, respectively.} \label{tab:ablation-components-continual}
\vskip -0.1in
\scalebox{0.9}{
\tabcolsep4pt
\begin{tabular}{l| cccccc|c || cccc|c}
\multicolumn{8}{c||}{a) continual} & \multicolumn{5}{c}{b) mixed domains} \\ \hline
Method  & \rotatebox[origin=c]{70}{CIFAR10-C} & \rotatebox[origin=c]{70}{CIFAR100-C} & \rotatebox[origin=c]{70}{ImageNet-C} & \rotatebox[origin=c]{70}{ImageNet-R} & \rotatebox[origin=c]{70}{ImageNet-Sk.} & \rotatebox[origin=c]{70}{ ImageNet-D109 } & Mean & \rotatebox[origin=c]{70}{CIFAR10-C} & \rotatebox[origin=c]{70}{ CIFAR100-C } & \rotatebox[origin=c]{70}{ImageNet-C} & \rotatebox[origin=c]{70}{ ImageNet-D109 } & Mean \\
\hline
  Source                    & 43.5 & 46.4 & 82.0 & 63.8 & 75.9 & 58.8 & 61.7 & 43.5 & 46.4 & 82.0 & 58.8 & 57.7 \\ 
  TENT                      & 20.0 & \textcolor{red}{62.2} & 62.6 & 57.6 & 69.5 & 52.9 & 54.1 &  \textcolor{red}{44.1} & \textcolor{red}{82.5} & \textcolor{red}{86.4} & 56.1 & \textcolor{red}{67.3} \\
 \hline
 SLR             & 20.1 & \textcolor{red}{57.7} & 61.5 & 55.6 & 67.8 & 52.7 & 52.6 & 42.8 & \textcolor{red}{78.2} & \textcolor{red}{87.4} & 58.2 & \textcolor{red}{66.7} \\
  + Loss weighting   & 17.7 & 31.1 & 60.8 & \textbf{51.1} & \textbf{64.1} & 52.0 & 46.1 & \textbf{26.9} & 35.2 & 72.1 & 51.6 & 46.4 \\
 + Weight ensembling & 17.7 & 29.5 & 56.2 & 52.3 & 65.5 & 48.9 & 45.0 & 29.1 & 35.4 & 71.4 & 51.5 & 46.9 \\
 + Consistency   & 16.3 & \textbf{29.3} & \textbf{54.4} & 51.2 & 64.2 & 48.1 & \textbf{43.9} & 28.4 & 35.1 & 69.6 & 51.0 & 46.0 \\
 + Prior correction & \textbf{16.2} & \textbf{29.3} & 54.5 & 51.2 & 64.3 & \textbf{48.0} & \textbf{43.9} & 28.0 & \textbf{35.0} & \textbf{69.5} & \textbf{50.9} & \textbf{45.9} \\
\hline
\end{tabular}}
\end{table}

\begin{table}[h]
\renewcommand{\arraystretch}{1.2}
\centering
\caption{Average online classification error rate~(\%) over 5 runs for different configurations for a) the \textit{correlated} TTA setting and b) the \textit{mixed domains correlated} TTA setting. For the ImageNet variants, a ViT-b-16 is used, while for CIFAR10-C a ResNet26-GN is applied.} \label{tab:ablation-components-correlated}
\vskip -0.1in
\scalebox{0.9}{
\tabcolsep4pt
\begin{tabular}{l| ccccc|c || ccc|c}
\multicolumn{7}{c||}{a) correlated} & \multicolumn{4}{c}{b) mixed + correlated} \\ \hline
Method  & \rotatebox[origin=c]{70}{CIFAR10-C} & \rotatebox[origin=c]{70}{ImageNet-C} & \rotatebox[origin=c]{70}{ImageNet-R} & \rotatebox[origin=c]{70}{ImageNet-Sk.} & \rotatebox[origin=c]{70}{ ImageNet-D109 } & Mean & \rotatebox[origin=c]{70}{CIFAR10-C} & \rotatebox[origin=c]{70}{ImageNet-C} & \rotatebox[origin=c]{70}{ ImageNet-D109 } & Mean \\
\hline
  Source                    & 32.7 & 60.2 & 56.0 & 70.6 & 53.6 & 54.6 & 32.7 & 60.2 & 53.6 & 48.8 \\ 
  TENT                      & \textcolor{red}{87.6} & \textcolor{red}{80.6} & 53.4 & 66.7 & \textcolor{red}{84.3} & \textcolor{red}{74.5} & \textcolor{red}{88.2} & \textcolor{red}{81.3} & \textcolor{red}{77.3} & \textcolor{red}{82.3} \\
 \hline
 SLR             & \textcolor{red}{89.0} & \textcolor{red}{90.3} & 52.3 & \textcolor{red}{78.0} & \textcolor{red}{90.4} & \textcolor{red}{80.0} & \textcolor{red}{88.3} & \textcolor{red}{88.7} & \textcolor{red}{87.4} & \textcolor{red}{88.1} \\
 + Loss weighting & 29.1 & \textcolor{red}{91.6} & 50.4 & 63.1 & \textcolor{red}{67.6} & \textcolor{red}{60.4} & \textcolor{red}{41.9} & \textcolor{red}{88.9} & 52.6 & \textcolor{red}{61.1} \\
 + Weight ensembling      & 28.1 & 44.7 & 49.8 & 61.7 & 49.2 & 46.7 & 31.0 & 53.9 & 49.8 & 44.9 \\
 + Consistency   & 29.5 & 42.5 & 48.0 & 60.5 & 48.1 & 45.7 & 31.0 & 51.5 & 48.8 & 43.8 \\
 + Prior correction & \textbf{15.9} & \textbf{16.8} & \textbf{25.8} & \textbf{44.0} & \textbf{31.7} & \textbf{26.8} & \textbf{17.4} & \textbf{23.6} & \textbf{29.4} & \textbf{23.5} \\
\hline
\end{tabular}}
\end{table}

\newpage
\section{Comparison to Related Work}
\paragraph{Comparison with CoTTA}
While both CoTTA and our proposed method utilize source weights, CoTTA uses stochastic restoring, where with a small probability current weights are restored with the corresponding weights from the source model. The idea behind stochastic restoring is that the network avoids drifting too far away from the initial source model. But, as discussed in Section \ref{sec:appendix_forgetting}, CoTTA first of all cannot prevent catastrophic forgetting on the continual ImageNet-C benchmark with 50,000 samples per corruption and, second, shows instabilities for certain domain shifts or settings. Instead of performing a stochastic restore, our proposed weight ensembling, which continually ensembles the weights of the initial source model and the weights of the current model, prevents catastrophic forgetting and mostly preserves the generalization capabilities of the initial source model.

\paragraph{Comparison with EATA} 
EATA, like our proposed method, utilizes certainty and diversity weighting. However, their weighting scheme relies on dataset-specific hyperparameters, such as an entropy threshold and a cosine similarity threshold. While the entropy threshold is determined heuristically, the cosine similarity threshold needs to be manually specified for each dataset. Choosing an inappropriate cosine similarity threshold can lead to a significant decrease in performance. For example, switching the cosine similarity threshold of CIFAR10-C and CIFAR100-C reduces the performance by absolutely 2.7\% and 10.8\%, respectively. In contrast, our proposed diversity weighting scheme does not necessitate dataset-specific hyperparameters and has demonstrated success across a wide range of different datasets, models, and domain shifts, as validated by our experiments. To address catastrophic forgetting, EATA incorporates elastic weight consolidation, which requires access to source samples for computing the Fisher information matrix. As discussed in \Cref{sec:appendix_forgetting}, our proposed weight ensembling approach also effectively mitigates catastrophic forgetting without the need for source data availability. Furthermore, EATA does not only exhibit instabilities when dealing with correlated data, but also demonstrates impractical performance outcomes in this setting due to not employing any prior correction.

\end{document}